\documentclass{article}

\usepackage{arxiv}

\usepackage[utf8]{inputenc} 
\usepackage[T1]{fontenc}    
\usepackage{hyperref}       
\usepackage{url}            
\usepackage{booktabs}       
\usepackage{amsfonts}       
\usepackage{nicefrac}       
\usepackage{microtype}      
\usepackage{lipsum}
\usepackage{graphicx}
\usepackage{subfig}

\title{Evolving Nano Particle Cancer Treatments with Multiple Particle Types}

\author{
    Michail-Antisthenis Tsompanas  \\
    Unconventional Computing Laboratory,\\
    University of the West of England,\\ 
    Bristol BS16 1QY, UK \\
    \texttt{antisthenis.tsompanas@uwe.ac.uk}
\And
       Larry Bull \\
     Department of Computer Science and \\Creative Technologies, \\ University of the West of England, \\ Bristol BS16 1QY, UK
\And
       Andrew Adamatzky \\
       Unconventional Computing Laboratory,\\
    University of the West of England,\\ 
    Bristol BS16 1QY, UK \\
\And
       Igor Balaz\\
       Laboratory for Meteorology, Physics and Biophysics,\\ Faculty of Agriculture, \\ Trg Dositeja Obradovica 8, \\University of Novi Sad, 21000, Novi Sad, Serbia
}

\begin{document}
\maketitle

\begin{abstract}
Evolutionary algorithms have long been used for optimization problems where the appropriate size of solutions is unclear a priori. The applicability of this methodology is here investigated on the problem of designing a nano-particle (NP) based drug delivery system targeting cancer tumours. Utilizing a treatment comprising of multiple types of NPs is expected to be more effective due to the higher complexity of the treatment. This paper begins by utilizing the well-known \textit{NK} model to explore the effects of fitness landscape ruggedness upon the evolution of genome length and, hence, solution complexity. The size of a novel sequence and the absence or presence of sequence deletion are also considered. Results show that whilst landscape ruggedness can alter the dynamics of the process, it does not hinder the evolution of genome length. These findings are then explored within the aforementioned real-world problem. In the first known instance, treatments with multiple types of NPs are used simultaneously, via an agent-based open source physics-based cell simulator. The results suggest that utilizing multiple types of NPs is more efficient when the solution space is explored with the evolutionary techniques under a predefined computational budget.
\end{abstract}


\section{Introduction}
\label{S:1}

Simulated evolution has long used variable-length representations with the aim of enabling the complexity of solutions to match those of the task faced. For instance, Fogel et al.'s \cite{fogel1965artificial} pioneering work included the use of mutation operators which could add or remove a node from finite state machines. A subset of a variable-length problems, established as metameric representation problems~\cite{ryerkerk2019survey}, can be solved by utilizing a segmented variable-length genome. This means that the solutions are defined as a set of similar components. Examples of these problems include the layout of wind farms, wireless sensor networks, and composite laminate stacking problems~\cite{ryerkerk2017solving}. Previously, several methodologies have been proposed, including alternative representations, mutation, recombination and selection operators, and these methodologies were compared among them~\cite{ryerkerk2019survey}. Variable-length algorithms have proven to be more efficient than the fixed-length ones in some cases, even if the optimal amount of components of the solution is known~\cite{ryerkerk2019survey}. This paper first explores the effects of fitness landscape ruggedness upon the evolution of genome length with an abstract tunable model and a mutation-based approach. The results are then investigated within an application in the bio-engineering domain, namely simulating a nano-particle (NP) based cancer treatment, where multiple different types of NPs can be included. 

In nature, novel sequences of DNA can originate through a variety of mechanisms, including retrotransposons, horizontal gene transfers, during recombination events, whole genome duplications and others. A novel sequence may have no immediate function and can be subsequently lost/selected due to a deleterious/beneficial mutation~\cite{ohno1970evolution}, may beneficially/detrimentally alter dosage~\cite{otto2000polyploid} or enable the subsequent specialisation of a duplicated function~\cite{hughes1994evolution}. Following~\cite{fogel1965artificial}, the process of novel variable sequence creation is here simplified such that a given number of random genes are added to an existing genome and immediately assigned random contributions to the organism’s fitness function. This is explored within the well-known \textit{NK} model \cite{kauffman1987towards} of fitness landscapes where size and ruggedness can be systematically altered and controlled. Results suggest that landscape ruggedness, the length of the new sequence with respect to that of the original genome, and the presence of gene deletion can all affect the evolution of genome length. Significantly, increases in genome length are seen across the parameter space of the model. 

These findings have been used to inform on-going investigations into optimizing the design of a simulated, NP based, targeted drug delivery system for cancer treatment. This simulation is based on PhysiCell \cite{ghaffarizadeh2018physicell} an open source physics-based cell simulator, which extends BioFVM \cite{ghaffarizadeh2015biofvm}, a large-scale transport solver. PhysiCell source code has been altered to simulate the injection of multiple types of NPs with different behaviours within the same treatment. However, the appropriate type of NPs for a given type of cancer tumour is unknown and, hence, optimization can be cast as a search through a space of variable size; a variable-length evolutionary algorithm is used to optimize both the number of types of NPs and their different features. It should be noted that, given the high complexity of the simulator and, thus, its long execution times, the evolutionary optimization is ideally kept under a computational budget. 

This paper presents initial findings of how the approach produces more complex treatments (more than one types of NPs) that are more effective than single NP treatments. Significantly, the number of NPs - or complexity - does not simply reach the maximum available. The average equilibrium complexity/size varies slightly depending upon the rate of increase in solution size used, the smaller the number of types of NPs added per mutation operation, the smaller the typical solution found. Since there is no significant difference in the fitnesses of both scenarios, for practical reasons (ease of manufacturing and lower toxicity), the solutions of smaller complexity can be considered as preferable.

The paper is arranged as follows: the next two sections present the NK model and results from it used to explore the underlying behaviour of a simple mutation-based approach to variable-length optimization. The following two sections present the results of its application to the bio-engineering task. Finally, the last section concludes the work.

\section{The NK Model}

\begin{figure}[!t]
    \centering
    \includegraphics[width=2.5in]{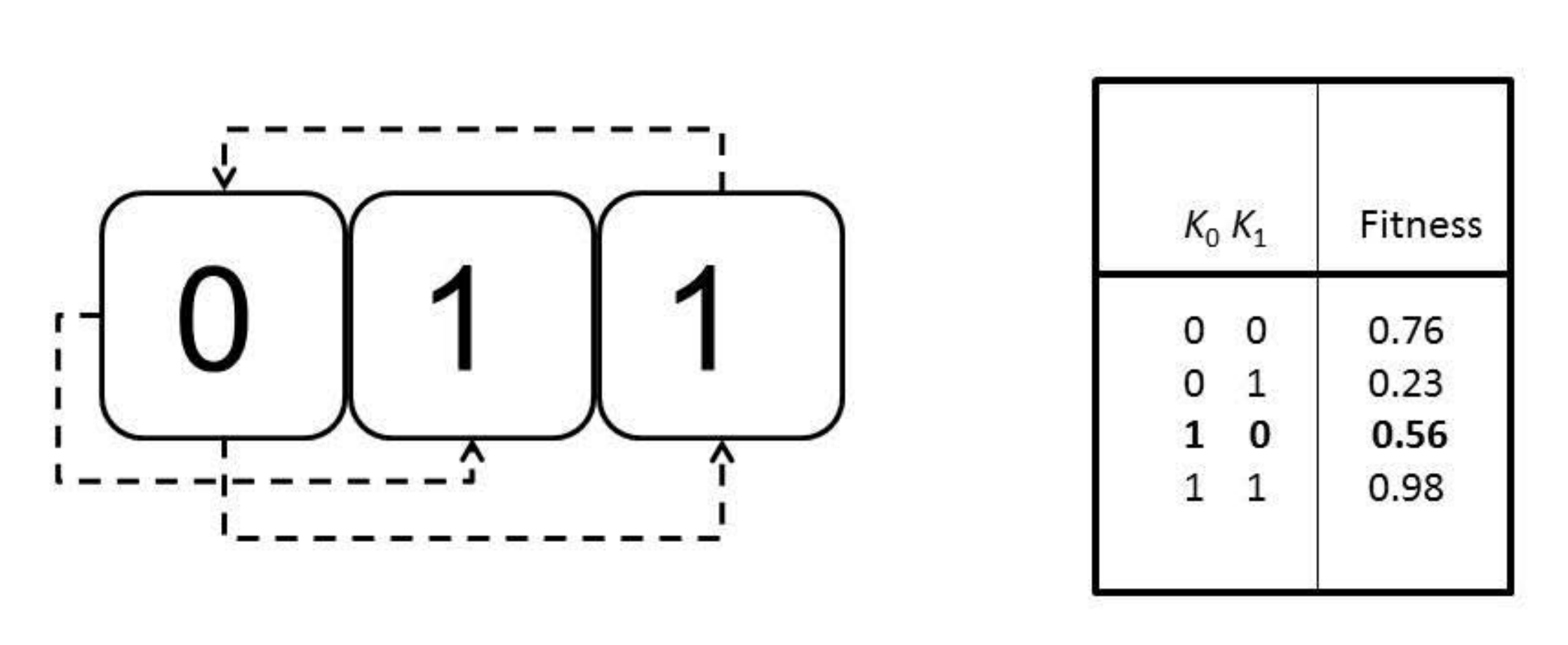}
    \caption{An example \textit{NK} model ($N=3$, $K=1$) showing how the fitness contribution of each gene depends on $K$ random genes (left). Therefore there are $2(K+1)$ possible allele combinations per gene, each of which is assigned a random fitness. Each gene of the genome has such a table created for it (right, centre gene shown). Total fitness is the normalized sum of these values.}
    \label{NKexample}
\end{figure}

Kauffman and Levin \cite{kauffman1987towards} introduced the \textit{NK} model to allow the systematic study of various aspects of fitness landscapes (see \cite{kauffman1993origins} for an overview). In the standard model, the features of the fitness landscapes are specified by two parameters: $N$, the length of the genome; and $K$, the number of genes that has an effect on the fitness contribution of each (binary) gene. Thus, increasing $K$ with respect to $N$ increases the epistatic linkage, increasing the ruggedness of the fitness landscape. The increase in epistasis increases the number of optima, increases the steepness of their sides, and decreases their correlation. The model assumes all intragenome interactions are so complex that it is only appropriate to assign random values to their effects on fitness. Therefore for each of the possible $K$ interactions a table of $2(K+1)$  fitnesses is created for each gene with all entries in the range 0.0 to 1.0, such that there is one fitness for each combination of traits (Fig. \ref{NKexample}). The fitness contribution of each gene is found from its table. These fitnesses are then summed and normalized by $N$ to give the selective fitness of the total genome.

\begin{figure*}[!tb]
    \centering
    \subfloat[]{\includegraphics[width=0.35\textwidth]{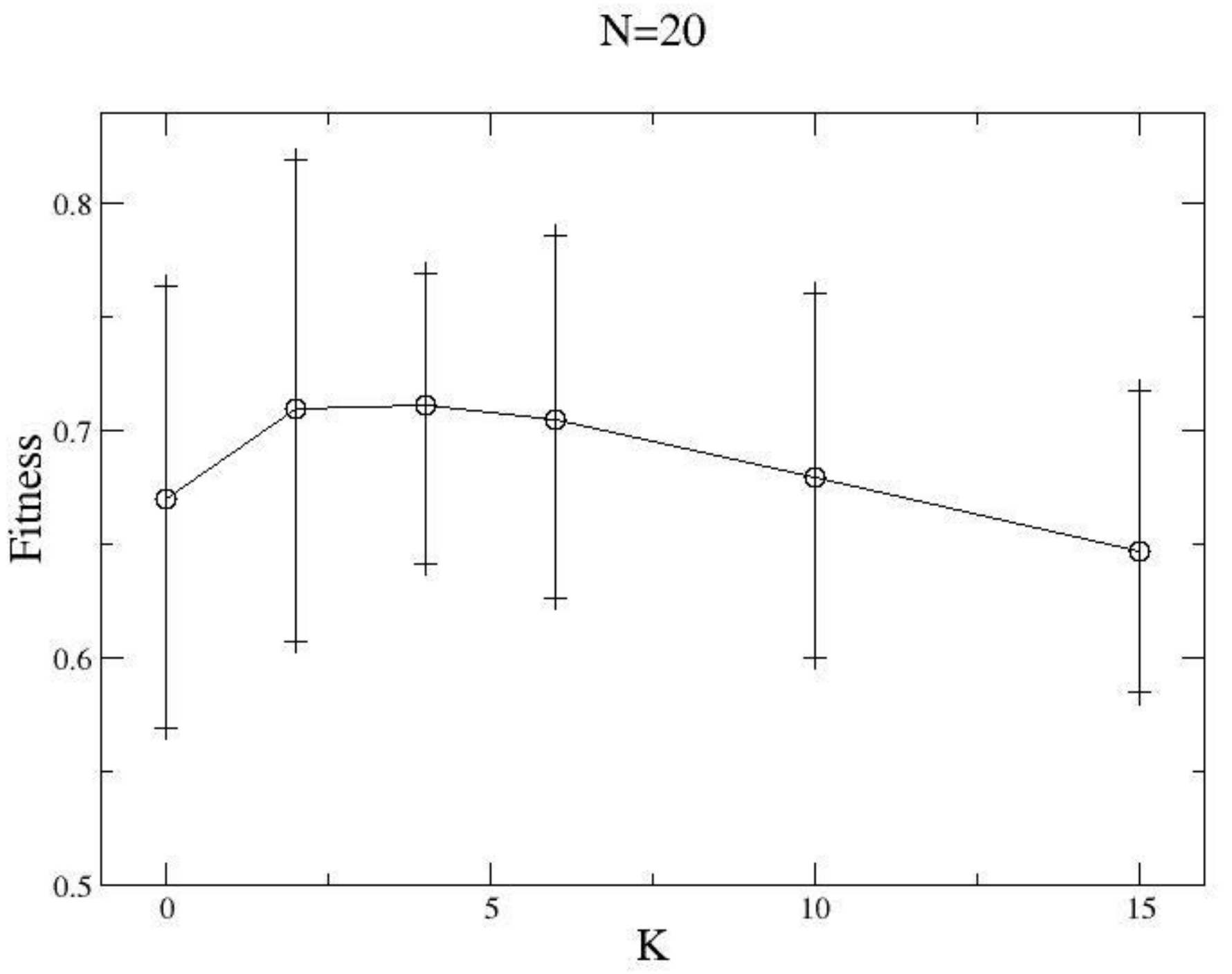}}
    \hfil
    \subfloat[]{\includegraphics[width=0.35\textwidth]{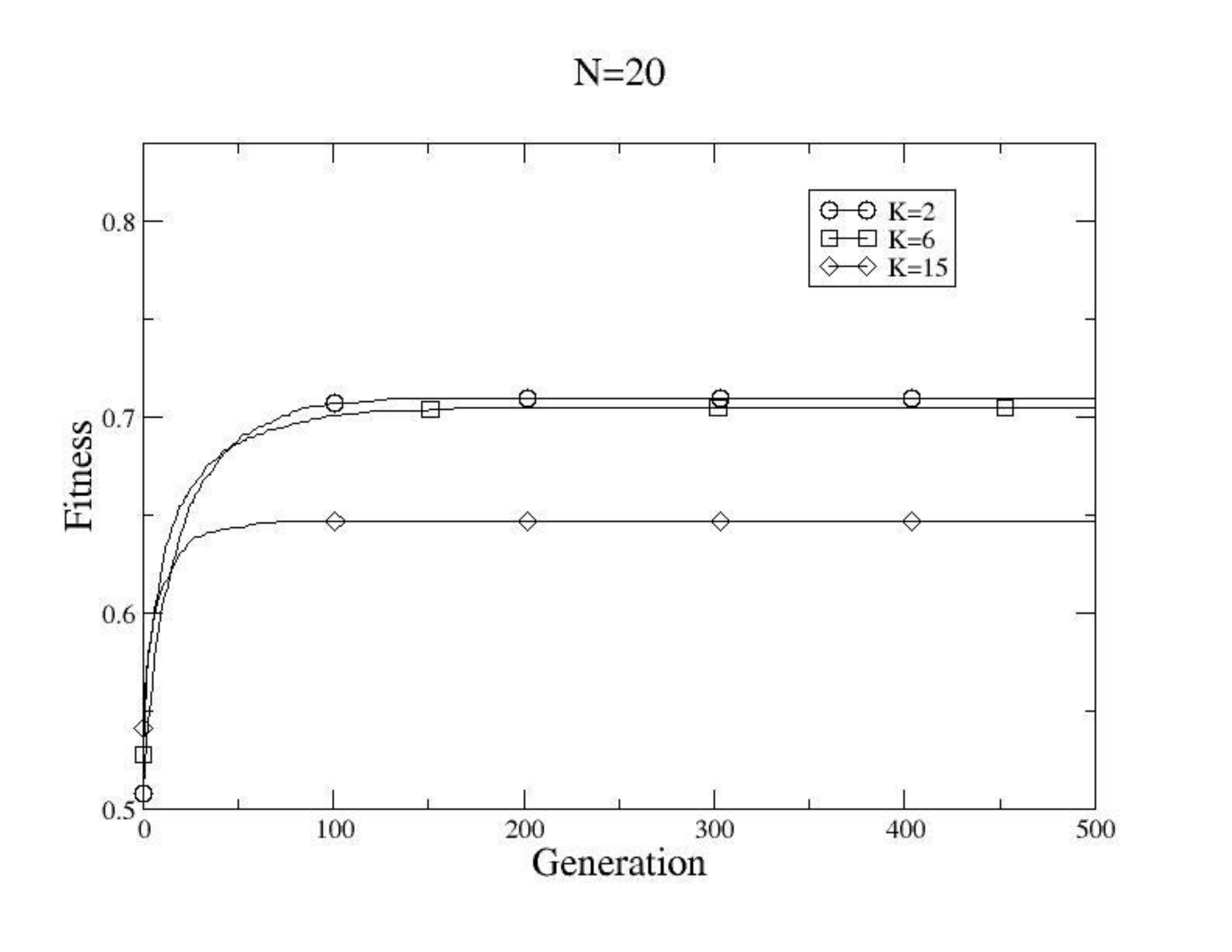}}
    \vfill
    \subfloat[]{\includegraphics[width=0.36\textwidth]{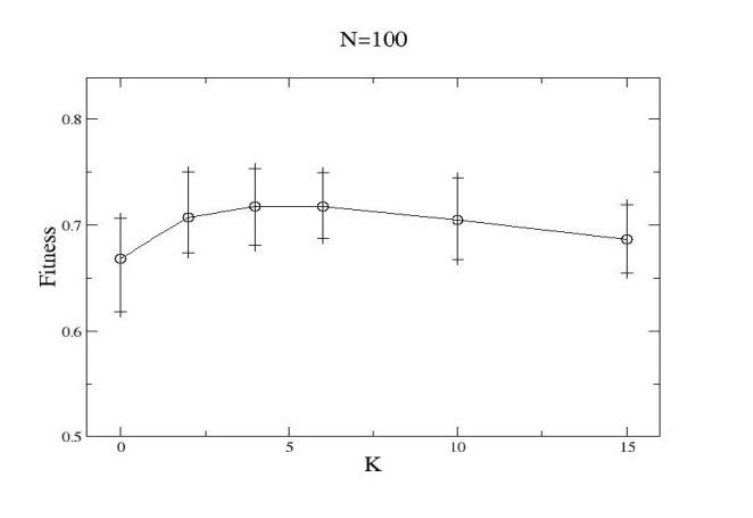}}
    \hfil
    \subfloat[]{\includegraphics[width=0.34\textwidth]{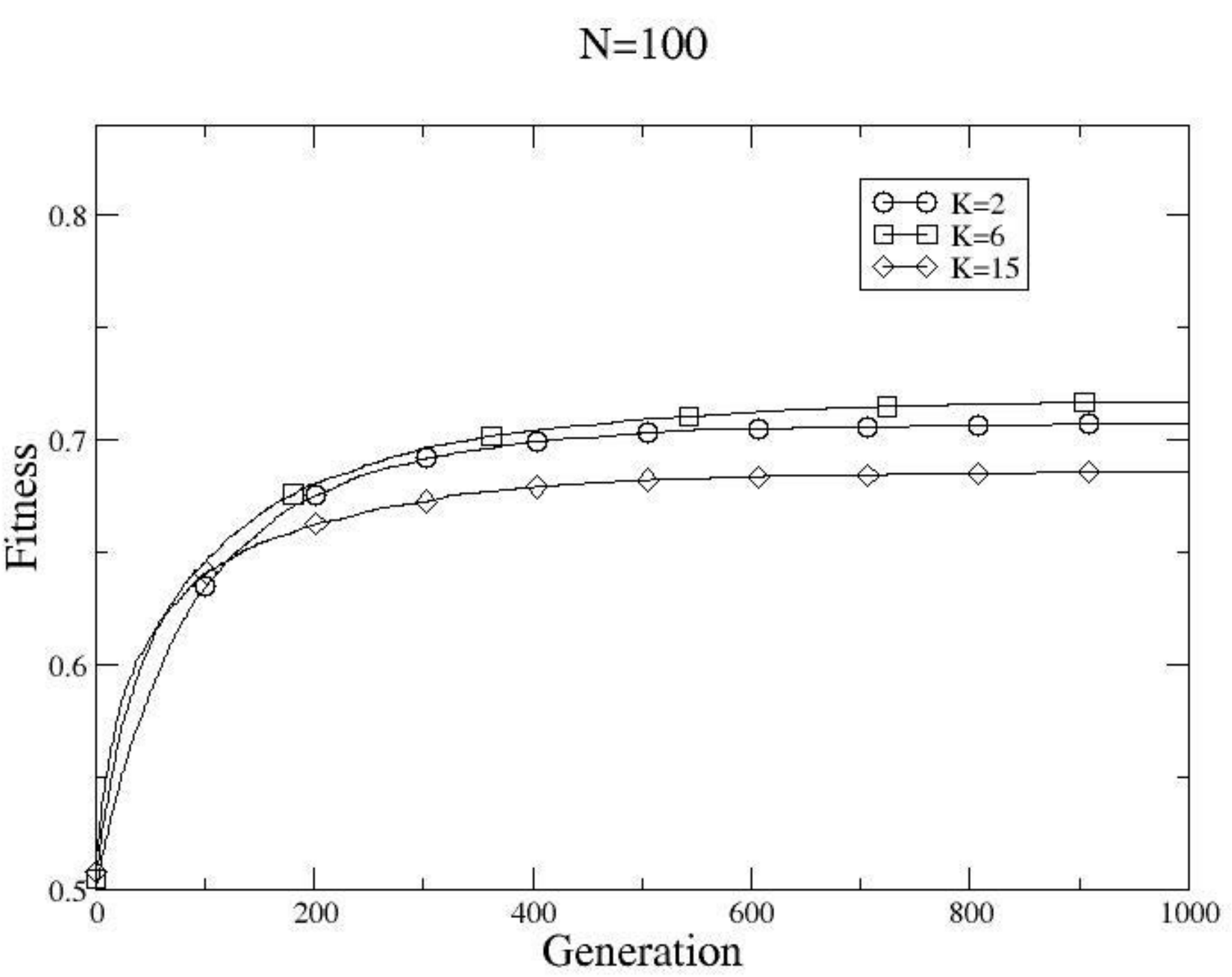}}
    \caption{Showing typical behaviour and the fitness reached after 20000 generations on landscapes of varying ruggedness ($K$) and length ($N$). Error bars show min and max values.}
    \label{typBehaviour}
\end{figure*}

Kauffman \cite{kauffman1993origins} used a mutation-based hill-climbing algorithm, where the single point in the fitness space is said to represent a converged species, to examine the properties and evolutionary dynamics of the \textit{NK} model. That is, the population is of size one and a species evolves by making a random change to one randomly chosen gene per generation. The ``population" is said to move to the genetic configuration of the mutated individual, if its fitness is greater than the fitness of the current individual; the rate of supply of mutants is seen as slow compared to the actions of selection. Ties are broken at random. Figure \ref{typBehaviour} shows example results. All results reported in this paper are the average of 10 runs (random start points) on each of 10 \textit{NK} functions, i.e., 100 runs, for 20000 generations. Here $0\leq K\leq 15$, for $N=20$ and $N=100$.


Figure \ref{typBehaviour} shows examples of the general properties of adaptation on such rugged fitness landscapes identified by Kauffman \cite{kauffman1993origins}, including a ``complexity catastrophe'' as $K\rightarrow N$. When $K=0$ all genes make an independent contribution to the overall fitness and, since fitness values are drawn at random between 0.0 and 1.0, order statistics show the average value of the fit allele should be 0.66. Hence a single, global optimum exists in the landscape of fitness 0.66, regardless of the value of $N$. At low levels of $K$ ($0<K<8$), the landscape buckles up and becomes more rugged, with an increasing number of peaks at higher fitness levels, regardless of $N$. Thereafter the increasing complexity of constraints between genes means the height of peaks typically found begin to fall as $K$ increases relative to $N$: for large $N$, the central limit theorem suggests reachable optima will have a mean fitness of 0.5 as $K\rightarrow N$. Figure \ref{typBehaviour} shows how the optima found when $K>6$ are significantly lower for $N=20$ compared to those for $N=100$ (T-test, $p<0.05$).

\section{Genome Growth in the NK Model}

\begin{figure*}[!t]
    \centering
    \subfloat[]{\includegraphics[width=0.35\textwidth]{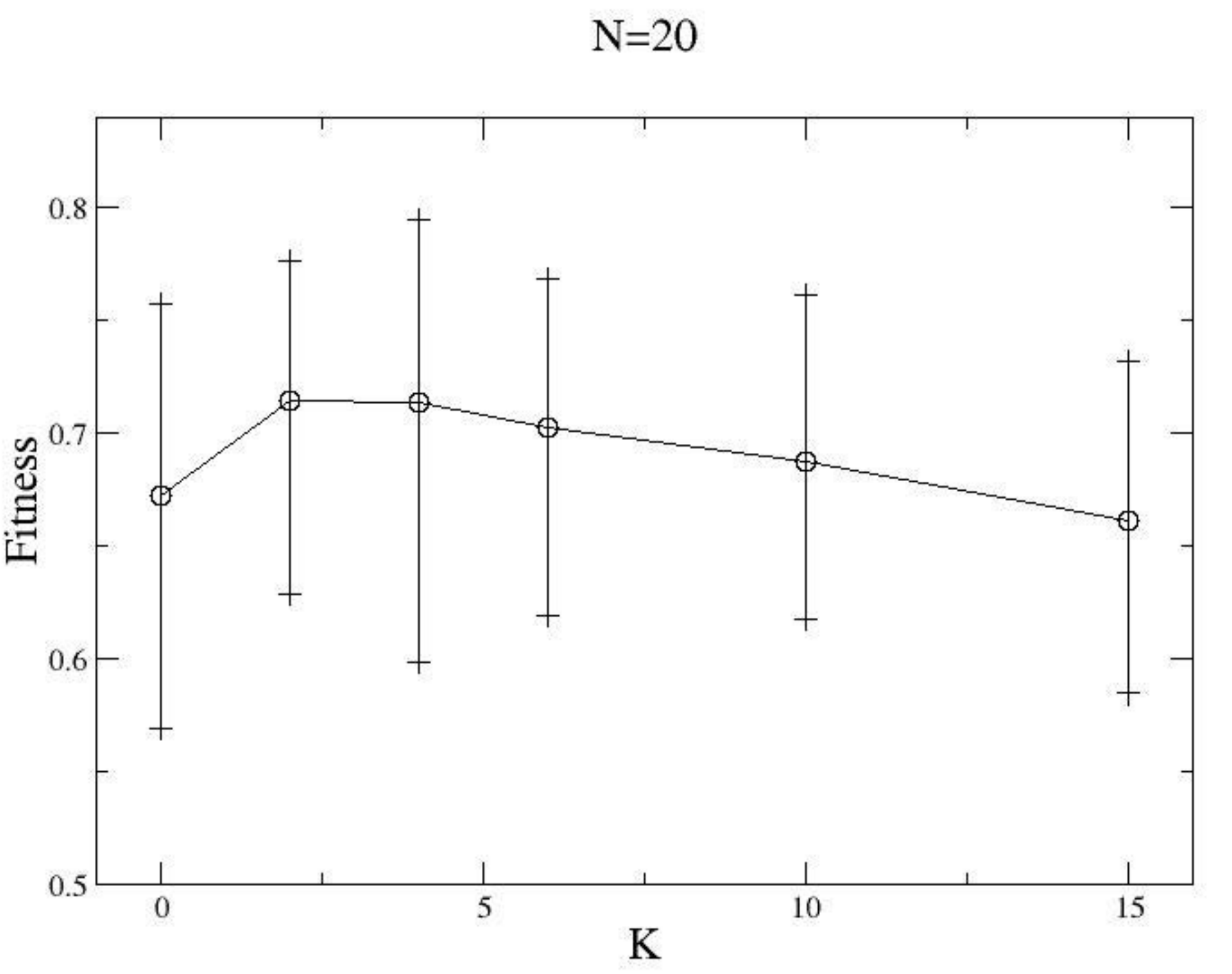}}
    \hfil
    \subfloat[]{\includegraphics[width=0.35\textwidth]{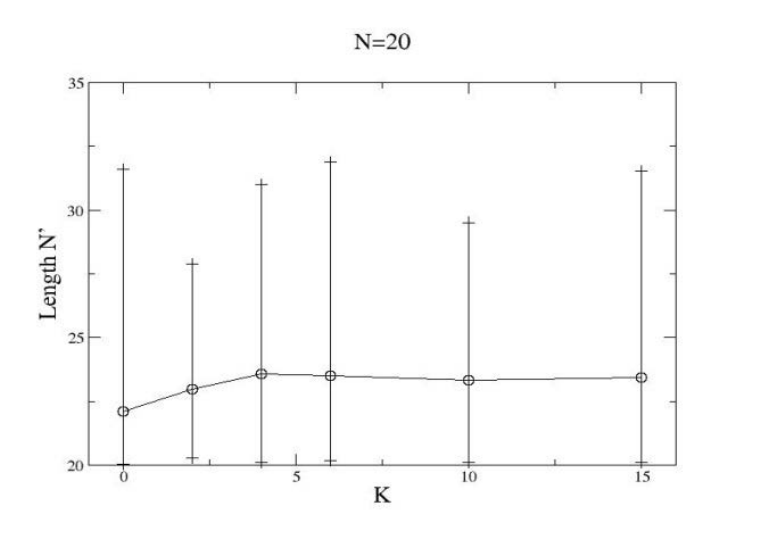}}
    \vfill
    \subfloat[]{\includegraphics[width=0.35\textwidth]{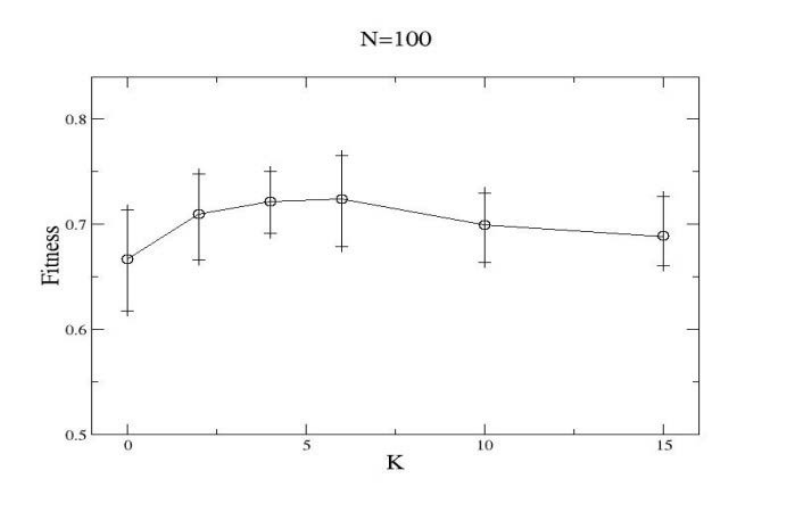}}
    \hfil
    \subfloat[]{\includegraphics[width=0.35\textwidth]{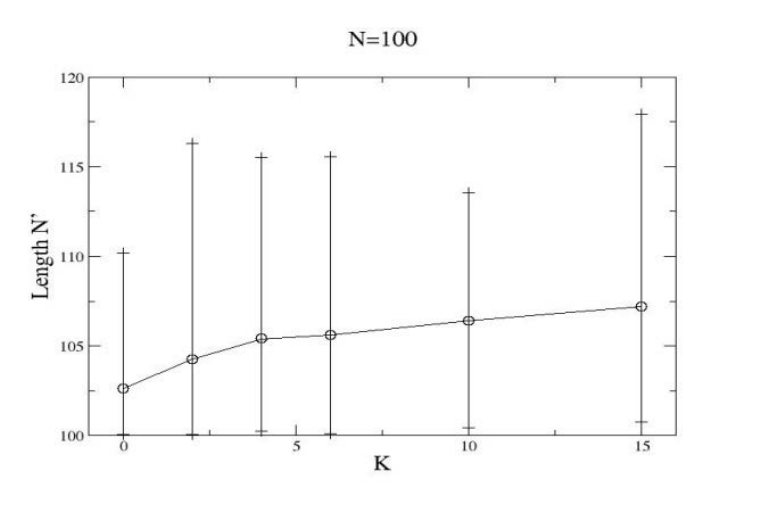}}
    \caption{Showing the fitness and length reached after 20000 generations on landscapes of varying ruggedness ($K$) where the initial length ($N$) can increase by one gene under mutation ($G=1$).}
    \label{fitLen}
\end{figure*}

\begin{figure*}[!t]
    \centering
    \subfloat[]{\includegraphics[width=0.35\textwidth]{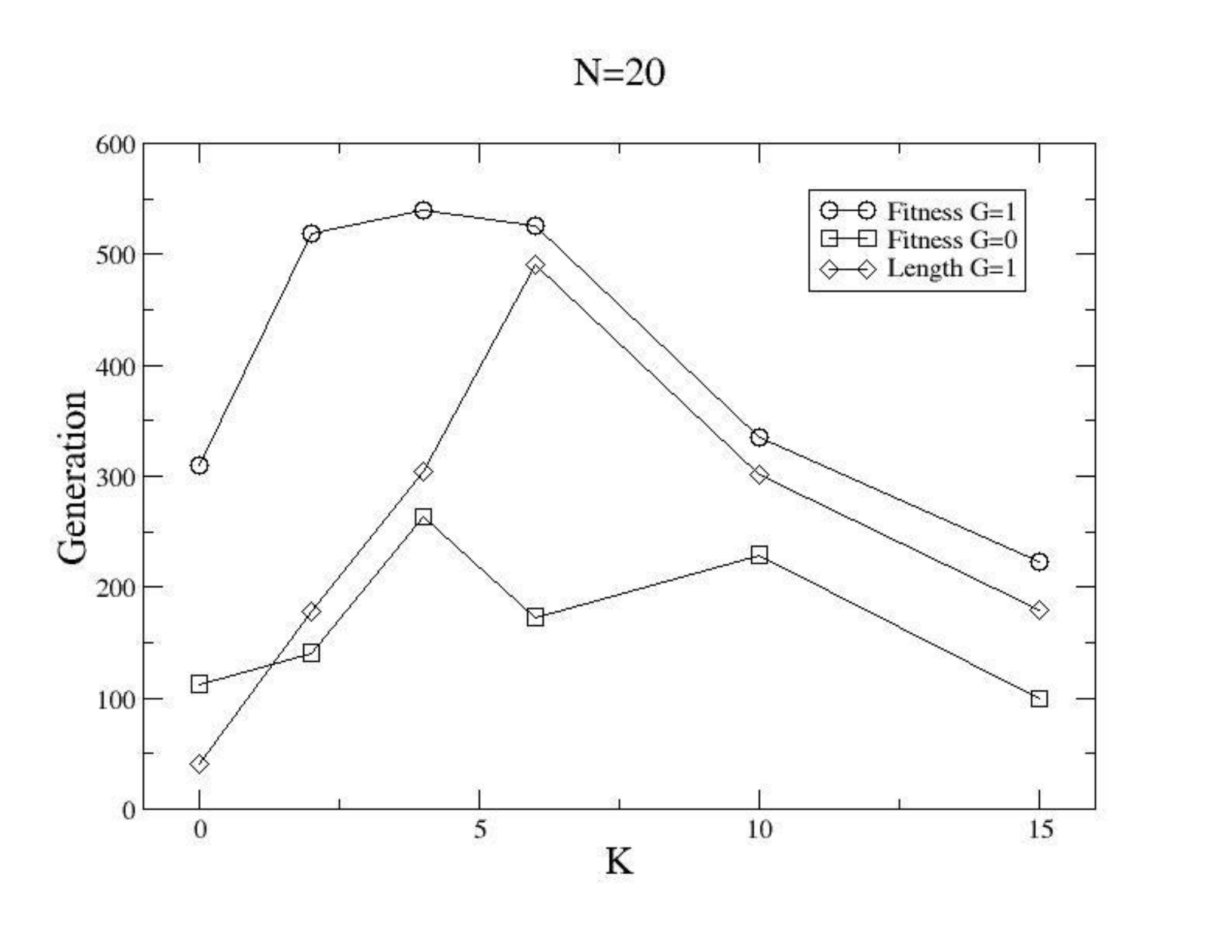}}
    \hfil
    \subfloat[]{\includegraphics[width=0.35\textwidth]{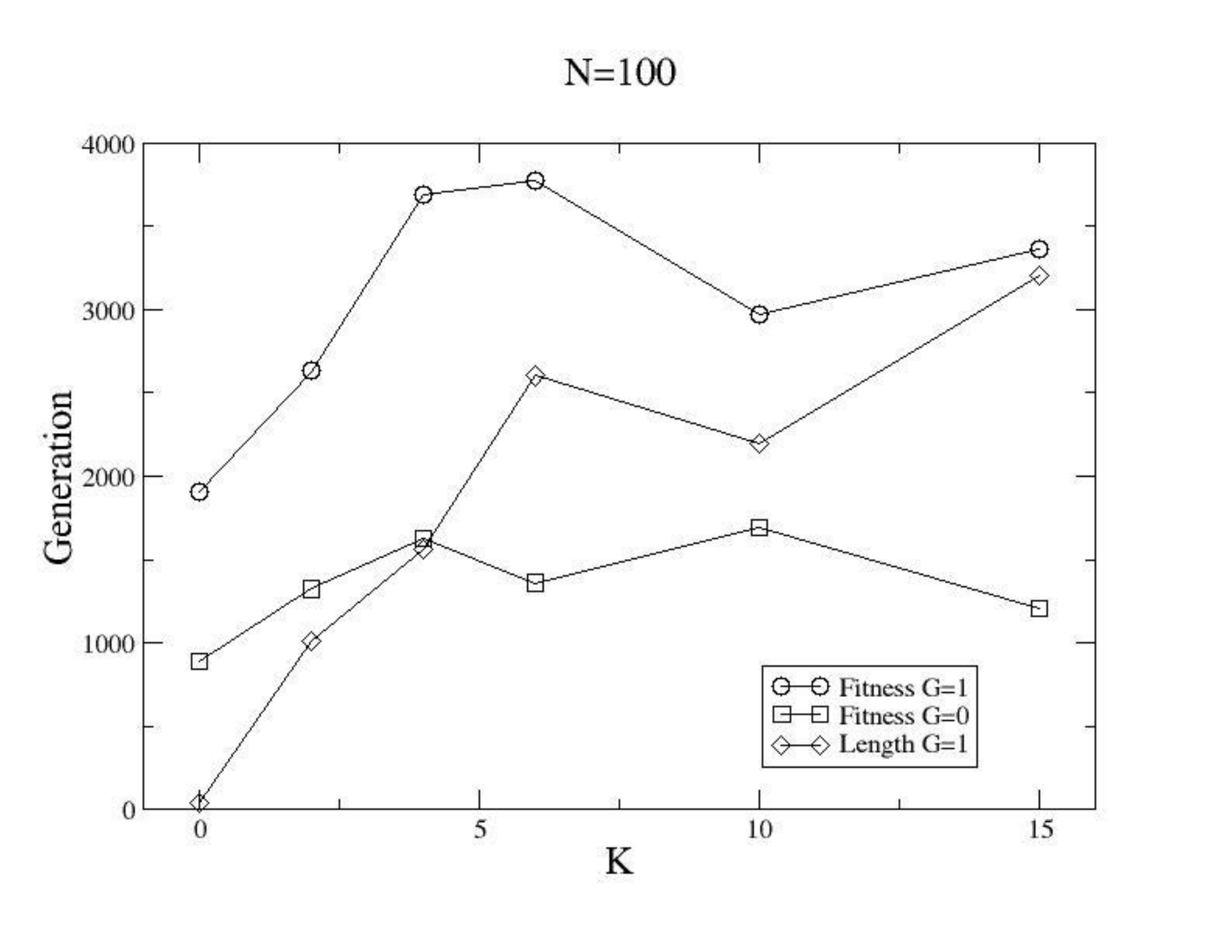}}
    \caption{Showing mean walk length to an optimum for a given $N$ and $K$, $G=1$}
    \label{meanWalk}
\end{figure*}

To enable the length of genomes in the \textit{NK} model to vary under evolution, i.e., to vary to $N’$, mutation is here expanded such that either a randomly chosen gene allele is altered as before or a number ($G$) of randomly created genes are added to the right-hand end of the existing genome. In the latter case, the first connection of a randomly chosen gene in the existing genome is assigned to each newly added gene (when $K\geq 1$). The new genes have $K$ randomly assigned connections into the whole genome, as before. In this way the new genes both affect and are affected by the existing genes.  

Figure \ref{fitLen} shows results from the simplest case of $G=1$. With a starting length $N=20$ the fitness increases for $K>10$ compared to the traditional case of $G=0$ (T-test, $p<0.05$). This is because the genomes typically increase in length by around 3 genes for all different tests with different values of $K$, i.e., $N’\approx 23$, which is sufficient to avert the onset of the complexity catastrophe at the highest $K$. There is a lot of variance in the genome lengths which emerge but slightly less growth is seen on average for $K<4$. When $N=100$ no significant change in fitness is seen for all K compared to $G=0$ (T-test, $p\geq 0.05$) but the typical amount of growth seen increases with $K$, where twice or more growth is seen for $K>4$ compared to the equivalent case with $N=20$.

\begin{figure*} [!t]
    \centering
    \subfloat[]{\includegraphics[width=0.35\textwidth]{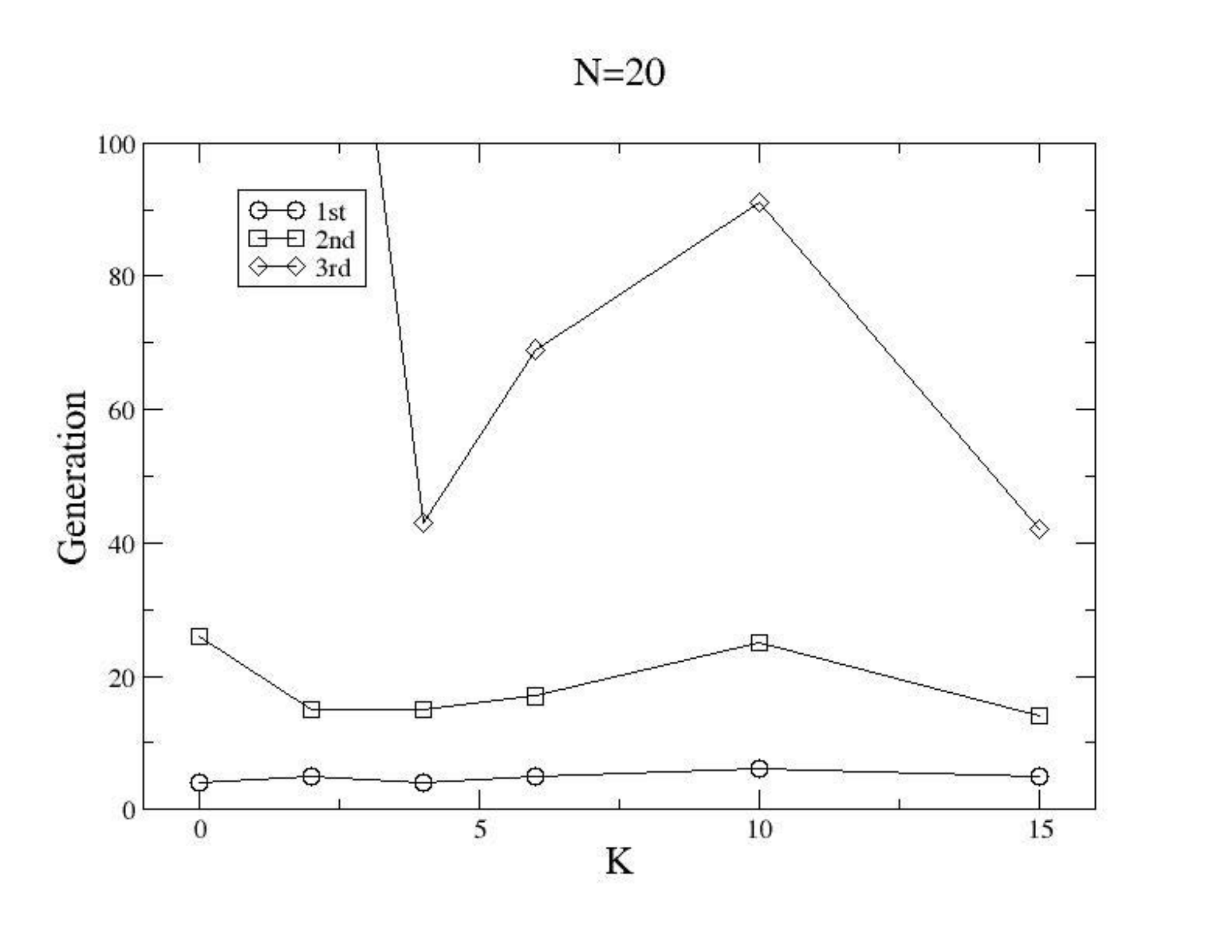}}
    \hfil
    \subfloat[]{\includegraphics[width=0.35\textwidth]{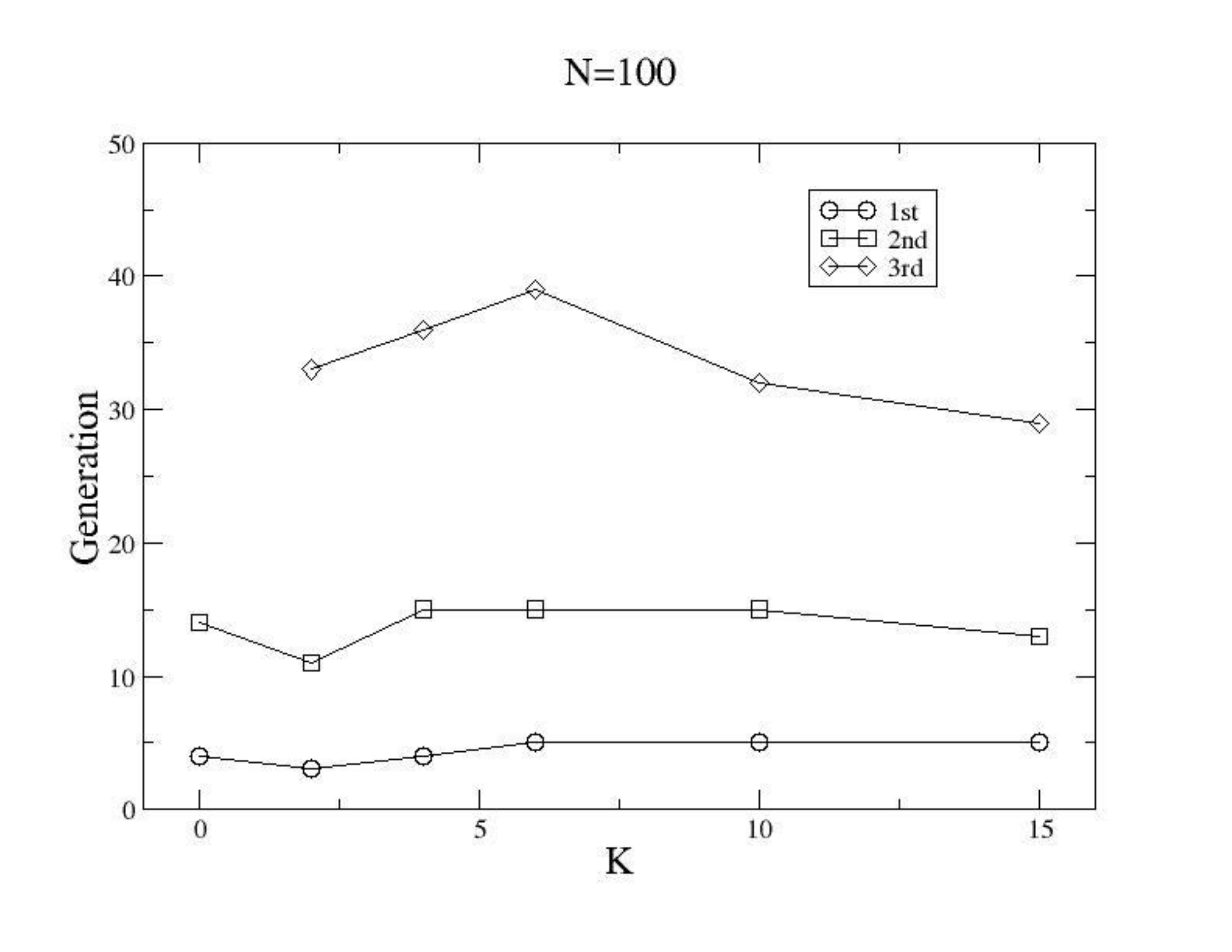}}
    \caption{Showing how the waiting time for each increase in length for a given N and K increases, $G=1$. Note $N=20$ and $K=2$ typically accepts its third gene at generation 179. Third genes are not added on average for $K=0$.}
    \label{waitTime}
\end{figure*}

As noted above, varying the degree of ruggedness varies the typical height and number of optima in a landscape. The more peaks of low fitness there are, the longer into the evolutionary search it is likely that a randomly added gene can make a beneficial contribution to fitness. Conversely, evolution can be anticipated to move through a series of relatively high fitness levels on correlated fitness landscapes in all but the earliest stages since fewer, higher peaks exist in the global space. Figure \ref{meanWalk} shows the generation at which evolution finds an optimum for various $K$, both with and without growth. It also shows the generation at which genome lengths stop changing. As can be seen, for all $K$, the addition of the genome length varying process means evolution continues for longer before finding an optimum – the dimensionality of the fitness landscape has increased. Moreover, it can also be seen that genome length stops increasing earlier for low $K$, continuing into the later stages of evolution as $K\rightarrow N$. Figure \ref{waitTime} shows the generation at which the first few new genes are typically accepted into the genome for various $K$. As can be seen, the waiting time does not vary significantly with $K$, i.e., landscape ruggedness, for the first two genes but begins to vary thereafter.

\begin{figure*}[!t]
    \centering
    \subfloat[]{\includegraphics[width=0.35\textwidth]{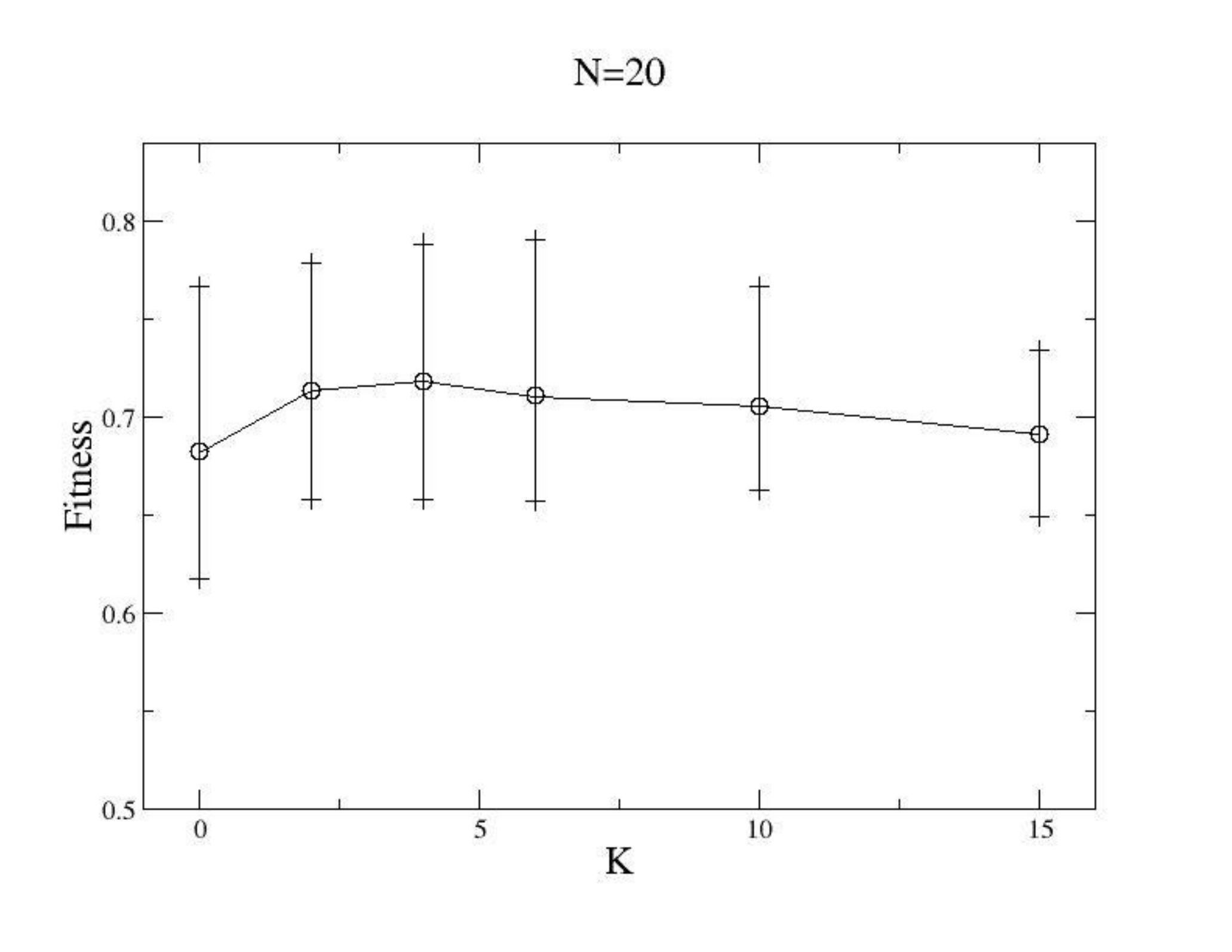}}
    \hfil
    \subfloat[]{\includegraphics[width=0.35\textwidth]{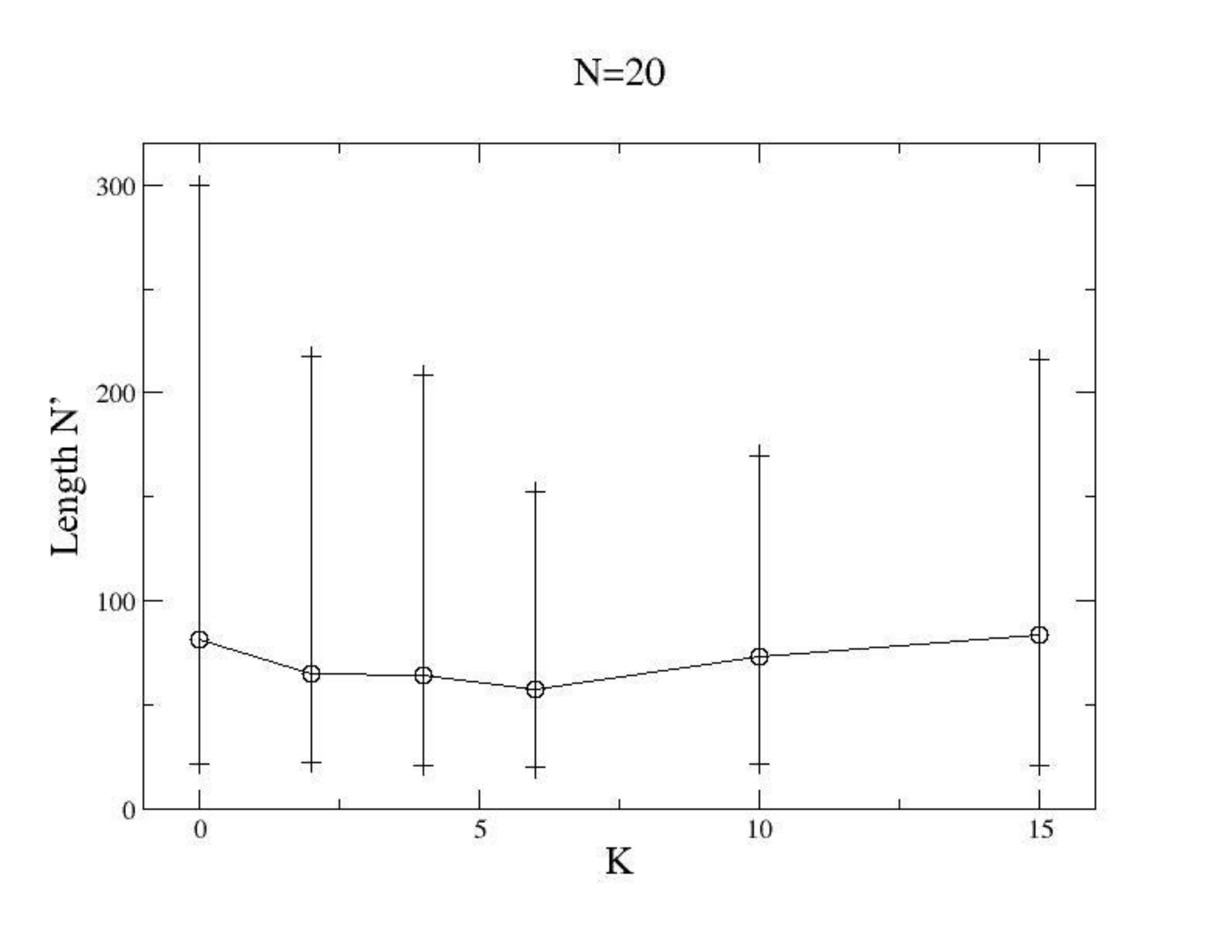}}
    \vfill
    \subfloat[]{\includegraphics[width=0.36\textwidth]{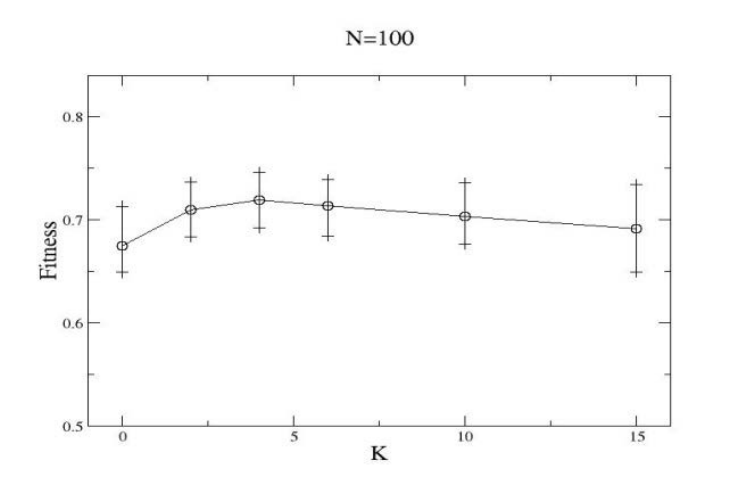}}
    \hfil
    \subfloat[]{\includegraphics[width=0.34\textwidth]{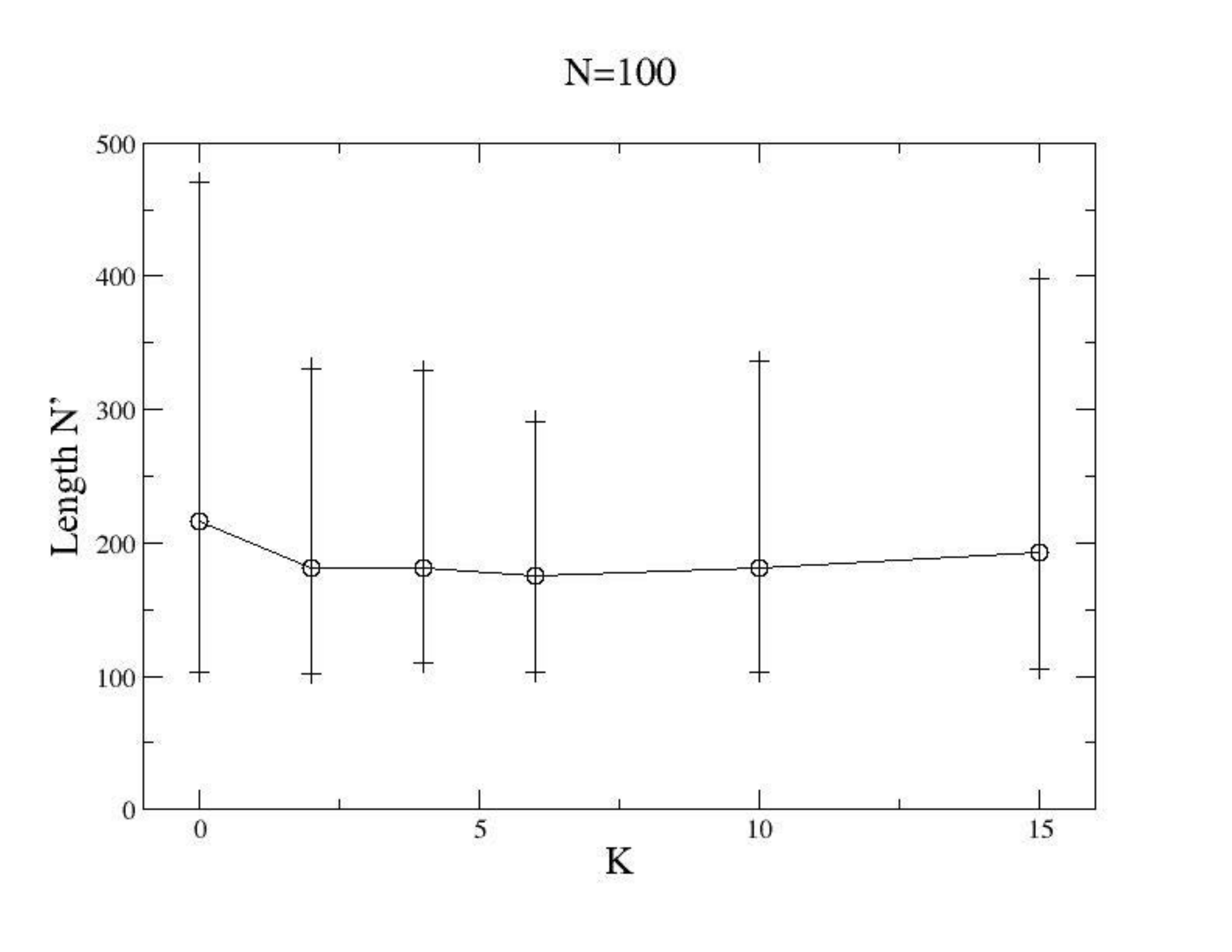}}
    \caption{Showing the fitness and length reached after 20000 generations on landscapes of varying ruggedness ($K$) where the initial length ($N$) can increase by twenty genes under mutation.}
    \label{fitLen2}
\end{figure*}

Figure \ref{fitLen2} shows the effect of increasing the size of the novel random sequence added, with $G=20$. When $N=20$, fitness is greater than with $G=1$ when $K>4$ (T-test, $p<0.05$) as the complexity catastrophe is further averted due to the significant increase in genome length for all $K$. When $N=100$, the fitness reached is the same as when $G=1$ for all $K$ with significantly longer genomes emerging.

\begin{figure*}[!t]
    \centering
    \subfloat[]{\includegraphics[width=0.35\textwidth]{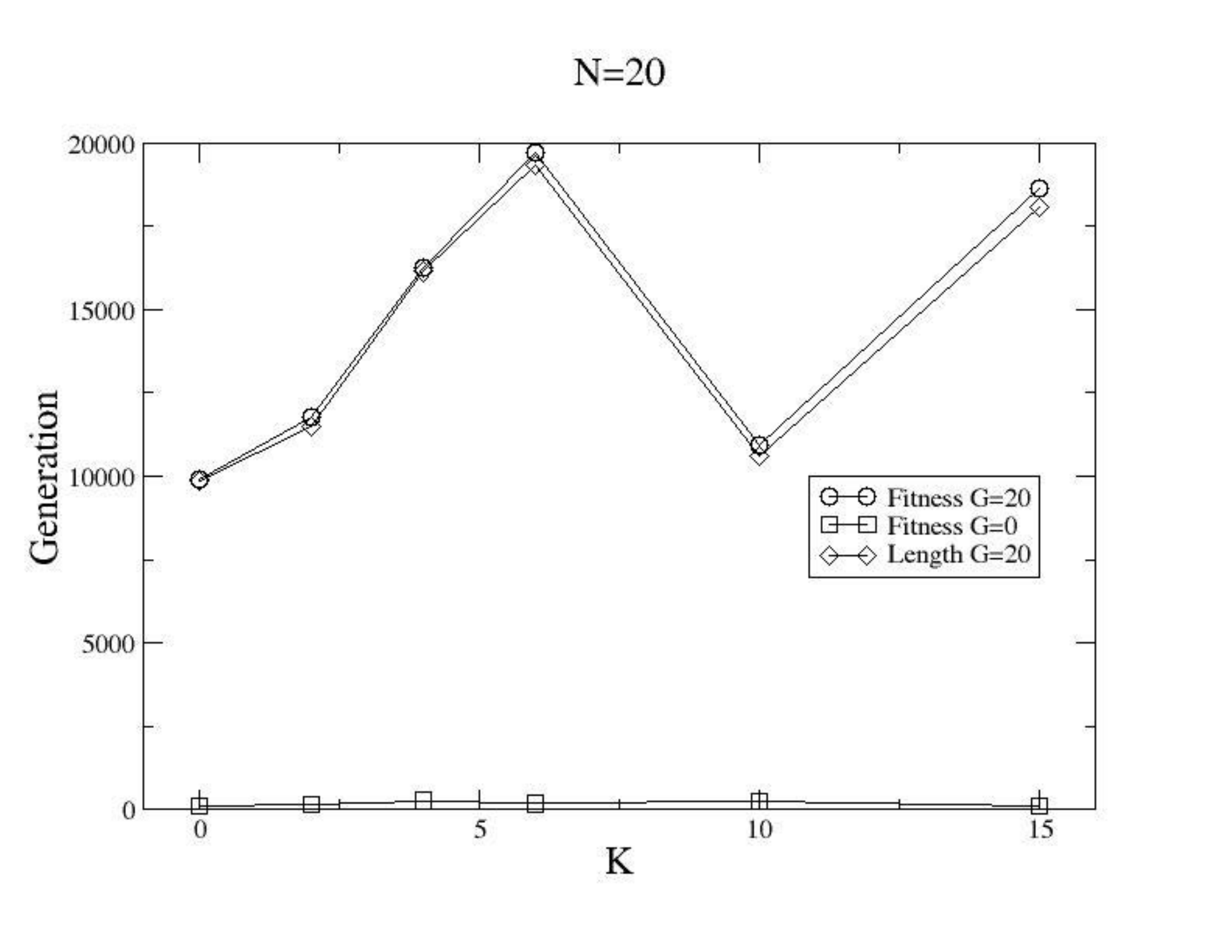}}
    \hfil
    \subfloat[]{\includegraphics[width=0.35\textwidth]{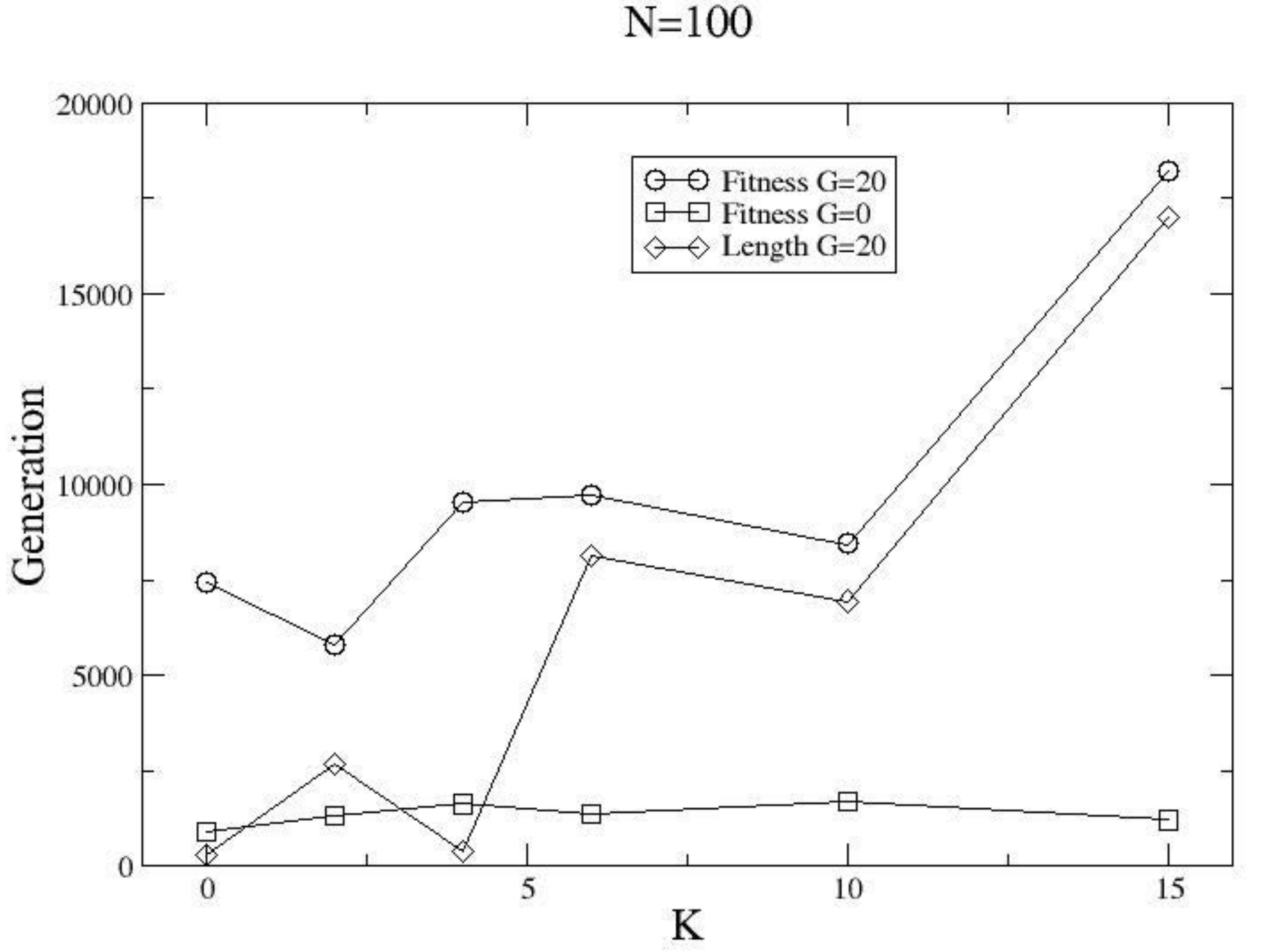}}
    \caption{Showing how mean walk length to an optimum for a given $N$ and $K$ increases with growth, $G=20$.}
    \label{meanWalk2}
\end{figure*}

\begin{figure*}[!t]
    \centering
    \subfloat[]{\includegraphics[width=0.35\textwidth]{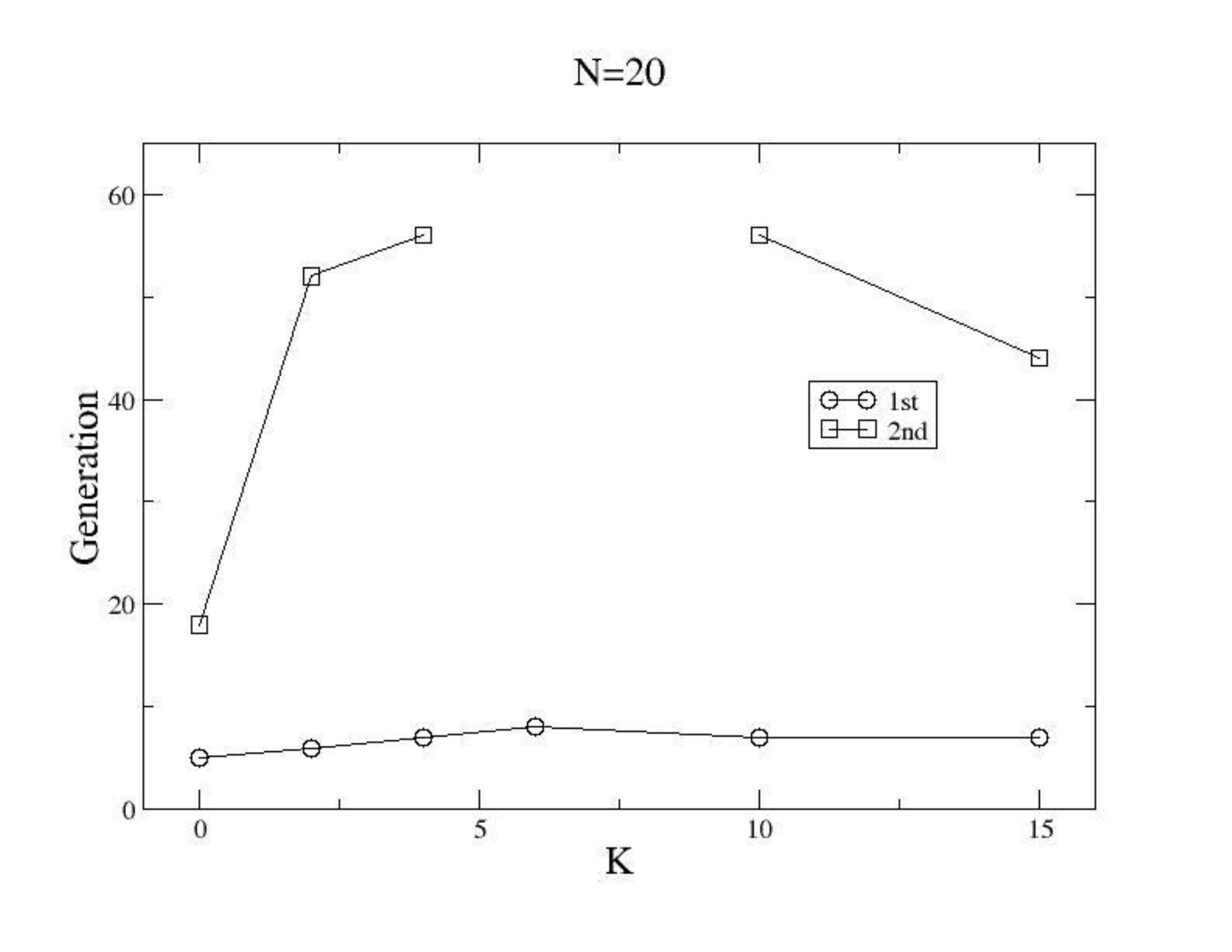}}
    \hfil
    \subfloat[]{\includegraphics[width=0.35\textwidth]{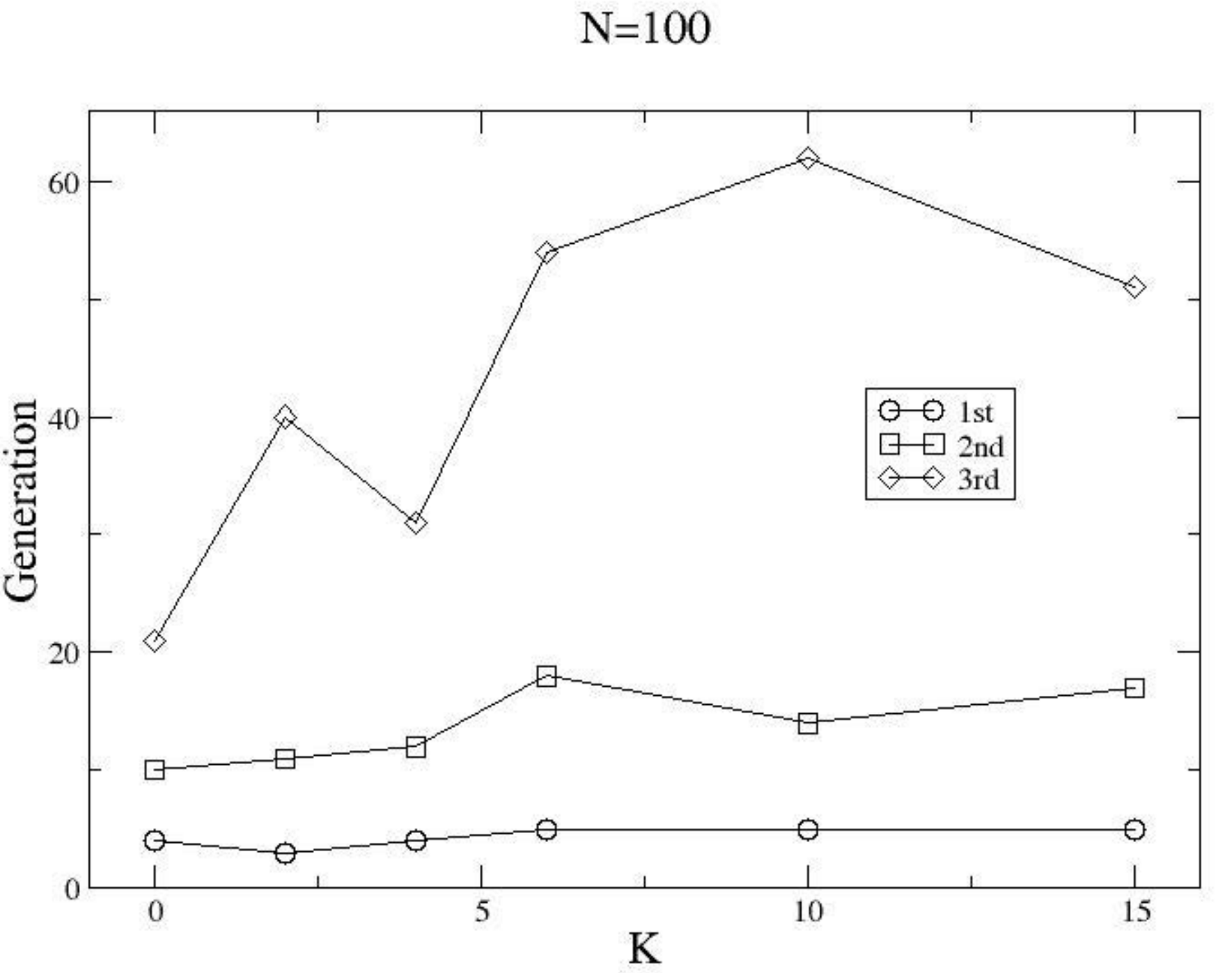}}
    \caption{Showing how the waiting time for each increase in length for a given $N$ and $K$ increases, $G=20$.}
    \label{waitTime2}
\end{figure*}

\begin{figure*}[!t]
    \centering
    \subfloat[]{\includegraphics[width=0.34\textwidth]{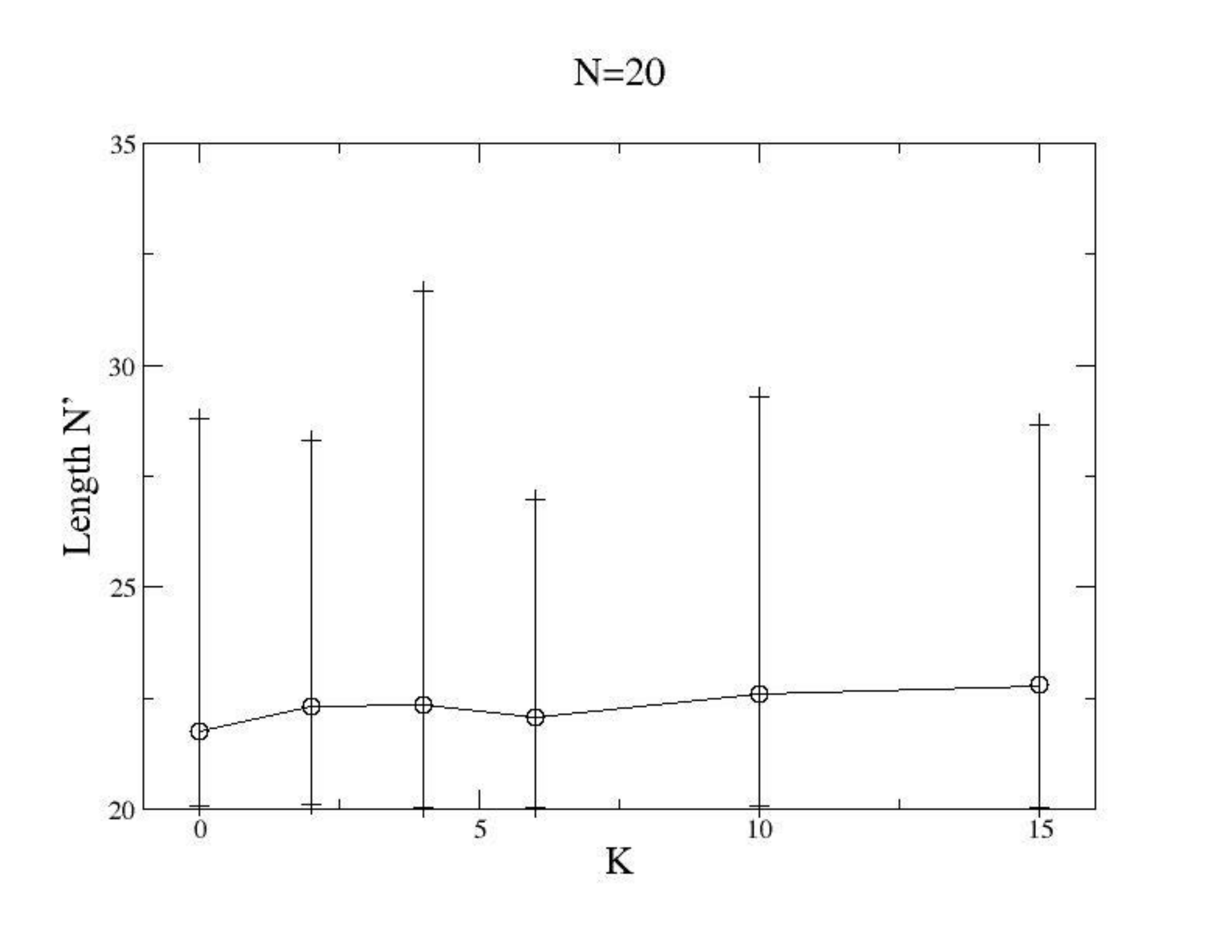}}
    \hfil
    \subfloat[]{\includegraphics[width=0.37\textwidth]{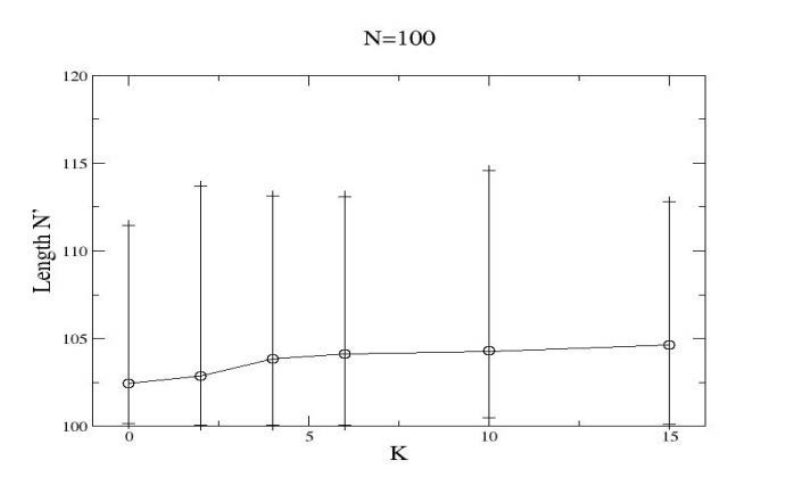}}
    \caption{Showing the lengths reached after 20000 generations on landscapes of varying ruggedness ($K$) where the initial length ($N$) can increase or decrease by one gene under mutation. }
    \label{lenRe}
\end{figure*}

Figures \ref{meanWalk2} and \ref{waitTime2} show the underlying dynamics of the evolutionary process in each case. As can be seen, the increased amount of growth per addition means evolution continues for much longer than with $G=1$ (Fig. \ref{meanWalk}), particularly when $N=20$. The general dynamics are not significantly different when $N=100$ but the relative increase in size of $G$ with respect to $N$ when $N=20$ is seemingly more disruptive. Figure \ref{waitTime2} shows how the first novel sequences are accepted at a similar generation as when $G=1$, regardless of $N$ and $K$. The second sequence is accepted at a similar stage when $N=100$ but at later stages for $N=20$, if at all. That is, the larger amount of growth per addition event can potentially maintain the conditions for more subsequent growth for longer; each increase in the dimensionality of a fitness landscape supplies a number of sub-optimal gene values, thereby maintaining sub-optimal fitness levels which in turn may aid the likelihood of a new random sequence being accepted.

\begin{figure*}[!t]
    \centering
    \subfloat[]{\includegraphics[width=0.35\textwidth]{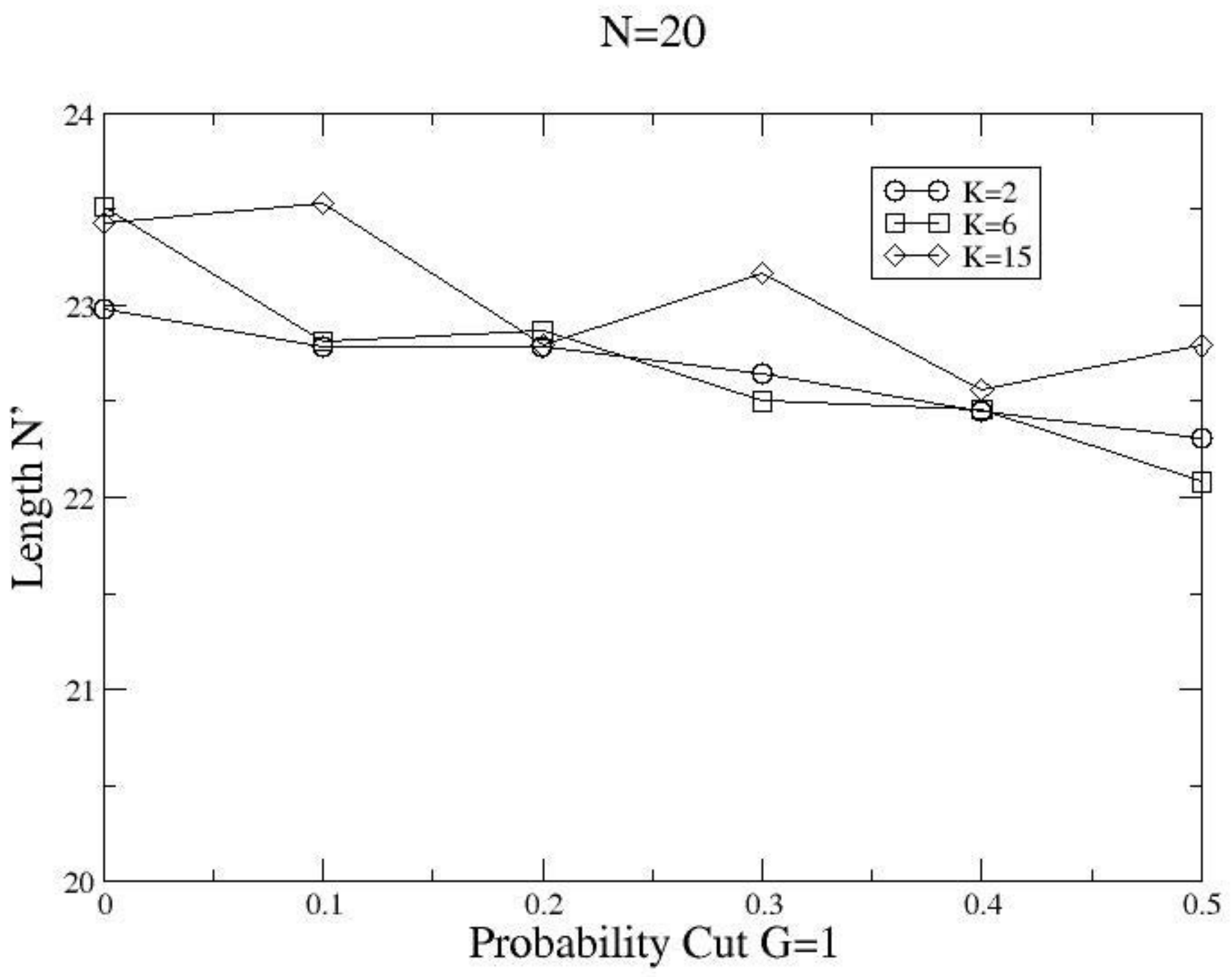}}
    \hfil
    \subfloat[]{\includegraphics[width=0.35\textwidth]{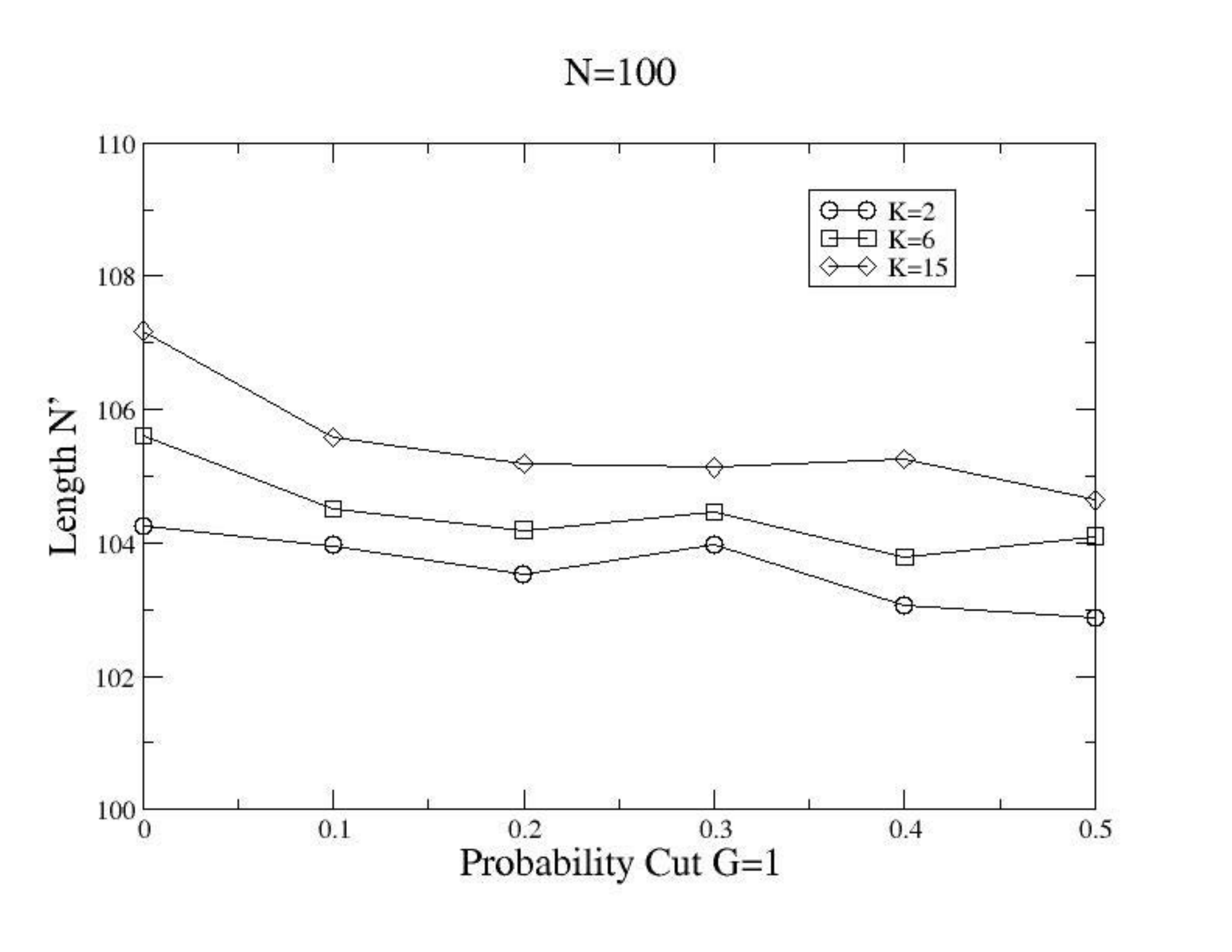}}
    \caption{Showing the effect of varying the probability of decreasing the number of genes for different $N$ and $K$.}
    \label{varEffect}
\end{figure*}

Using a similarly extended version of the \textit{NK} model, and $N=16$, $K=2$ (only), Harvey \cite{harvey1992species} showed that gradual growth through small increases in genome length were sustainable, whereas larger increases per growth event were not. This is explained as being due to the fact that a degree of correlation between the smaller fitness landscape and the larger one must be maintained; a fit solution in the former space must achieve a suitable level of fitness in the latter to survive into succeeding generations. Kauffman and Levin \cite{kauffman1987towards} discussed this general concept with respect to fixed-size \textit{NK} landscapes and varying mutation step sizes therein. They showed how for long jump adaptations, i.e., mutation steps of a size which go beyond the correlation length of a given fitness landscape, the time taken to find fitter variants doubles per generation. Harvey \cite{harvey1992species} draws a direct analogy between the size of the novel sequence being added ($G$) and the length of a jump in a traditional landscape; the larger $G$, the less correlated the two landscapes. He similarly points out that growth is more likely early in evolution before optima are climbed and it is after that the degree of correlation begins to take more effect. It is here suggested that his larger $G$ did not prove successful primarily due to a change in the standard \textit{NK} fitness function used to include length, and hence change in the selection pressure, which reduced the size of the window of opportunity for larger increases in length to emerge. Results, here, show larger $G$ increases of genome length are sustainable, for all $K$, i.e., regardless of the underlying correlation of the landscape. However, it can be seen in Fig. \ref{waitTime2} that when $N=20$ a second successful adoption of a novel sequence does not typically emerge when $4<K<10$ and a third adoption is rare (contrast with $G=1$ or $N=100$ and $G=20$). It may be that the degree of correlation between the original and new space is playing a role, although growth occurs for $K=2$ and $K=4$. It might be expected, although not explored here, that since genome lengths vary for almost as long as fitness levels when a larger $G$ is used, smaller amounts of growth would be accepted more readily during this time. That is, occasional large changes interspersed with many smaller changes, i.e., a dynamic value for G, would provide even more growth.

A common outcome for an added novel sequence in nature is removal, through mutational inactivation, fractionation after whole genome duplications, etc. Figure \ref{lenRe} shows example results from adding a process of deletion to balance that of growth described above. That is, an offspring can experience a gene allele mutation or genome length mutation with equal probability, as before. However, in the latter case, there is an equal probability that the last $G$ genes added are removed as for $G$ new genes to be added. Namely, a 50\% probability of gene mutation, a 25\% probability of adding and a 25\% probability of deleting a sequence. Fitness is unaffected for all $N$ and $K$ (not shown) but there is less growth for $K>0$. The same was also the case for $G=20$ (not shown).

Given the reduction, the sensitivity of the growth process to the relative rate of deletion to addition has also been explored. Figure \ref{varEffect} shows examples of the evolved lengths over the range of no deletion to an equal amount, with $G=1$. As can be seen, the rate of decline in genome length is roughly proportional to the rate of increase in the probability of deletion.

These findings are now used to inform explorations of novel types of NPs cancer treatment where the number of types is not pre-determined.

\section{PhysiCell}

The high computational capacity of modern systems have enabled the incorporation of detailed mathematical models in the study of biological processes, along with laboratory experiments. More specifically, the study of cancer biology is the epicenter of many mathematical models \cite{metzcar2019review} that aim to be powerful tools in the hands of scientists. One of these models is PhysiCell \cite{ghaffarizadeh2018physicell}, which is an agent-based, multicellular, open source simulator emulating physics and biological rules in interactions between cells. Moreover, PhysiCell utilizes BioFVM \cite{ghaffarizadeh2015biofvm} to emulate the secretion, diffusion and uptake of chemical substances in the simulated area.

Employing PhysiCell as a target simulator for optimization has been previously proposed in problems of designing NP-based drug delivery systems \cite{PREEN20191,tsompanas2019,tsompanas2020utilizing,tsompanas2020novelty} or unveiling cancer imunotherapies \cite{ozik2019learning}. The design of NPs was investigated through surrogate-assisted \cite{PREEN20191}, haploid-diploid \cite{tsompanas2019}, differential evolution \cite{tsompanas2020utilizing} and novelty search \cite{tsompanas2020novelty} evolutionary algorithms, while the most effective immunotherapy for cancer tumours was examined through a combination of active learning and genetic algorithms \cite{ozik2019learning}.

In specific, the sample project ``anti-cancer biorobots'' (see \cite{ghaffarizadeh2018physicell} for details) of PhysiCell simulator (v.1.6.1) was used to simulate the NPs interaction with the tumour, as in previous works \cite{PREEN20191,tsompanas2019,tsompanas2020novelty,tsompanas2020utilizing}. However, here the need to simulate complex multi-NP-based treatments led to an alternation in the source code. Nonetheless, similar to aforementioned works, here the design of a NP is defined as a point in the 5-dimensional space of the following parameters (with their range in the brackets): attached worker migration bias [0,1], unattached worker migration bias [0,1], worker relative adhesion [0,10], worker relative repulsion [0,10], worker motility persistence time (min) [0,10]. The rest of the user-defined parameters are not altered from the original instance published by the developers of the simulator and are presented in Table \ref{tabl:1}.

\begin{table}[!t]
\centering
\caption{Unaltered parameters of PhysiCell simulator.}
\label{tabl:1}       
\begin{tabular}{|l|l|}
\noalign{\smallskip}\hline
Parameter & Value  \\\hline
\noalign{\smallskip}\noalign{\smallskip}\hline
Damage rate &  0.03333 $min^{-1}$\\ \hline
Repair rate &  0.004167 $min^{-1}$\\ \hline
Drug death rate &  0.004167 $min^{-1}$\\ \hline
Elastic coefficient &  0.05 $min^{-1}$  \\ \hline
Cargo $O_2$ relative uptake   & 0.1 $min^{-1}$ \\ \hline
Cargo apoptosis rate       & 4.065e-5 $min^{-1}$ \\  \hline
Cargo relative adhesion    & 0  \\ \hline
Cargo relative repulsion   & 5  \\ \hline
Cargo release $O_2$ threshold & 10 $mmHg$ \\ \hline
Maximum relative cell adhesion distance & 1.25 \\  \hline
Maximum elastic displacement & 50 $\mu m$  \\ \hline
Maximum attachment distance & 18 $\mu m$ \\ \hline
Minimum attachment distance & 14 $\mu m$ \\  \hline
Motility shutdown detection threshold &  0.001 \\ \hline
Attachment receptor threshold  & 0.1 \\ \hline
Worker migration speed     & 2 $\mu m / min$ \\  \hline
Worker apoptosis rate      & 0 $min^{-1}$ \\  \hline
Worker $O_2$ relative uptake  & 0.1 $min^{-1}$ \\ \hline
\noalign{\smallskip}
\end{tabular}
\end{table}

The scenario in the sample project ``anti-cancer biorobots'' of agent-based PhysiCell simulator is the following (for more extensive details refer to \cite{ghaffarizadeh2018physicell}). An initial tumour with total radius of 200 $\mu m$ of cancer cells (approximately 570 cancer cell agents) is undergoing growth of a simulated period of 7 days. At that point the treatment is injected in the simulated area, comprising from 450 cargo agents and 50 worker agents, emulating therapeutic compound and NPs, respectively. A simulated period of 3 days is then executed, where the worker agents transport the cargo agents (therapeutic compound) and deposit them near the cancer cell agents that decay and die because of the affinity to the cargo cells. After the total simulated period of 10 days, the remaining amount of cancer cell agents in the simulated area is considered as the fitness function of the given point in the parameter space, describing the design of worker agents or simulated NPs.

Every simulation of these 10 days, was run on an Intel\textregistered{}  Xeon\textregistered{} CPU E5-2650 at 2.20GHz with 64GB RAM (using 8 of the 48 cores) and was completed at approximately 5 minutes of wall-clock time. Because of the stochastic nature of the simulator, a static sampling approach was employed, considering the mean of 5 runs with the same parameters. In order to further minimize the noise caused by the stochasticity of the simulator and, also, speed up the optimization process, alternations in the original source code were made in order to load a single tumour at the initialization of the simulator and apply the therapy for 3 days (as described in \cite{tsompanas2019}). 

One specific tumour, which was derived after a simulated growth of 7 days of an initial tumour with total radius of 200 $\mu m$, is used as the new initial state and the treatment is injected to the simulated area at $t=0$. Then, 3 days are simulated and, thus, the evaluation of a single solution needs 1,5 minutes (and for the 5 runs for static sampling 7,5 minutes). Each test in the following, starts with the same randomly produced population of solutions and evolves for 200 evaluations of each individual/solution. As a result, for each test the results are provided after approximately a day of continuously running the simulator.

Nonetheless, appropriate alternations in the source code of the simulator were made to accommodate the functionality of injecting the region around the cancer tumour with more than one type of NPs with different parameters. Without loss of generality a maximum of 10 different types of NPs was defined. The amount of NPs (worker agents) are set to 50, despite the amount of types of NPs. These 50 NPs are equally divided within the different types of each solution. For instance, with 2 types of NPs tests, 25 of each one are used; with 5 types of NPs, 10 of each one, etc.

\section{Genome growth in the simulator}

In this paper, the optimization of the design of NPs for drug delivery initially uses a standard steady-state genetic algorithm (GA) and mutation operators only for a single NP. The population size is $P = 20$, reproduction selection is implemented as a tournament of size $T = 2$, and replacement selection is the inverse. Offspring only replace the selected individual if its fitness is higher. The mutation procedure was executed on one randomly selected gene and modified with random step size of $s = [-5; 5]\%$. The computational budget was set as 1000 evaluations with PhysiCell, thus, given the 5 run static sampling approach, the population is evolved for 200 generations.

\begin{figure}[!t]
    \centering
    \includegraphics[width=0.45\textwidth]{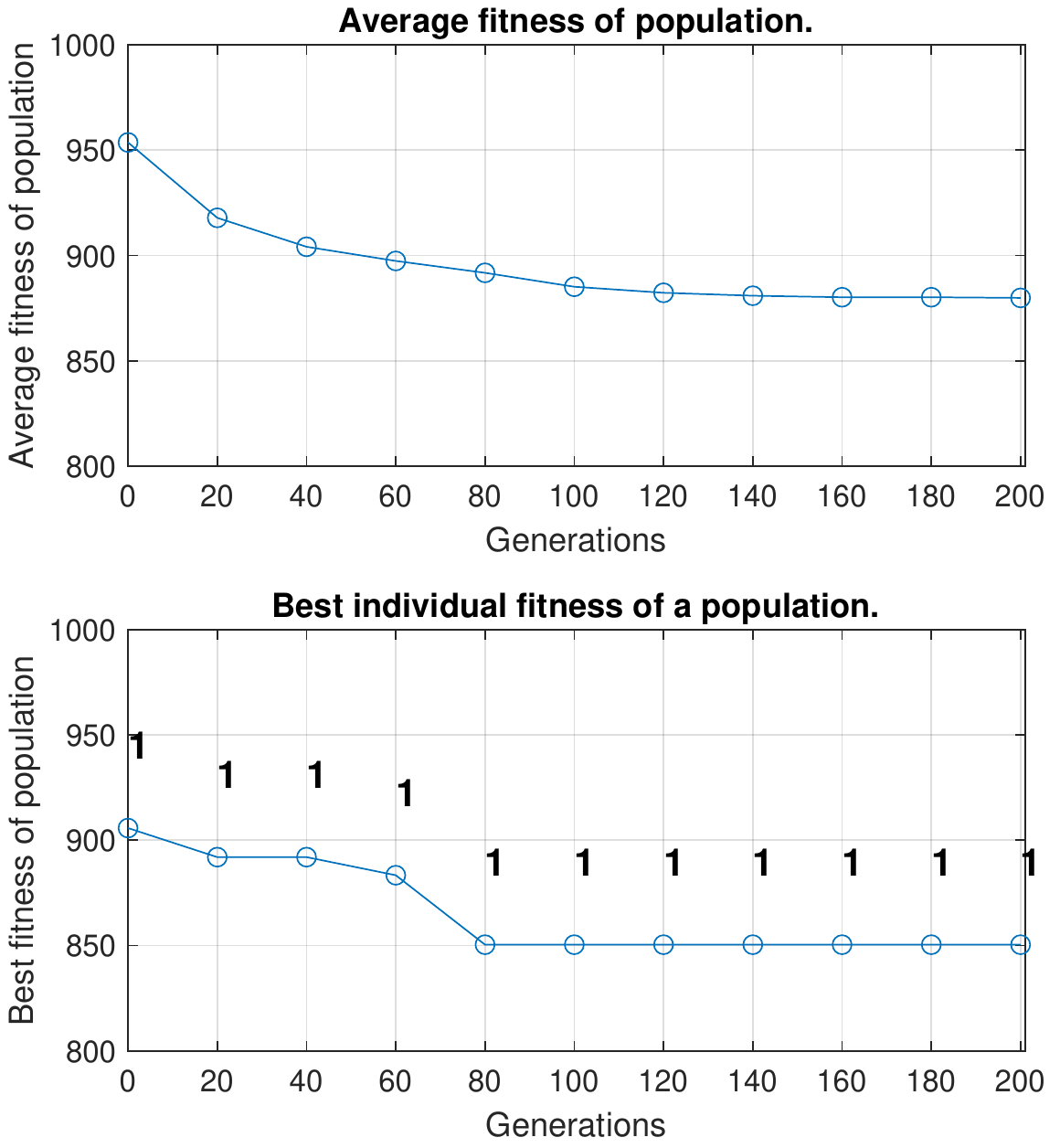}
    \caption{Results from one run of GA. (a) Evolution of average fitness of the population and (b) evolution of the best individual in the population (where numbers indicate the number of types NPs).}
    \label{physiRes1}
\end{figure}

\begin{figure}[!t]
    \centering
    \includegraphics[width=0.45\textwidth]{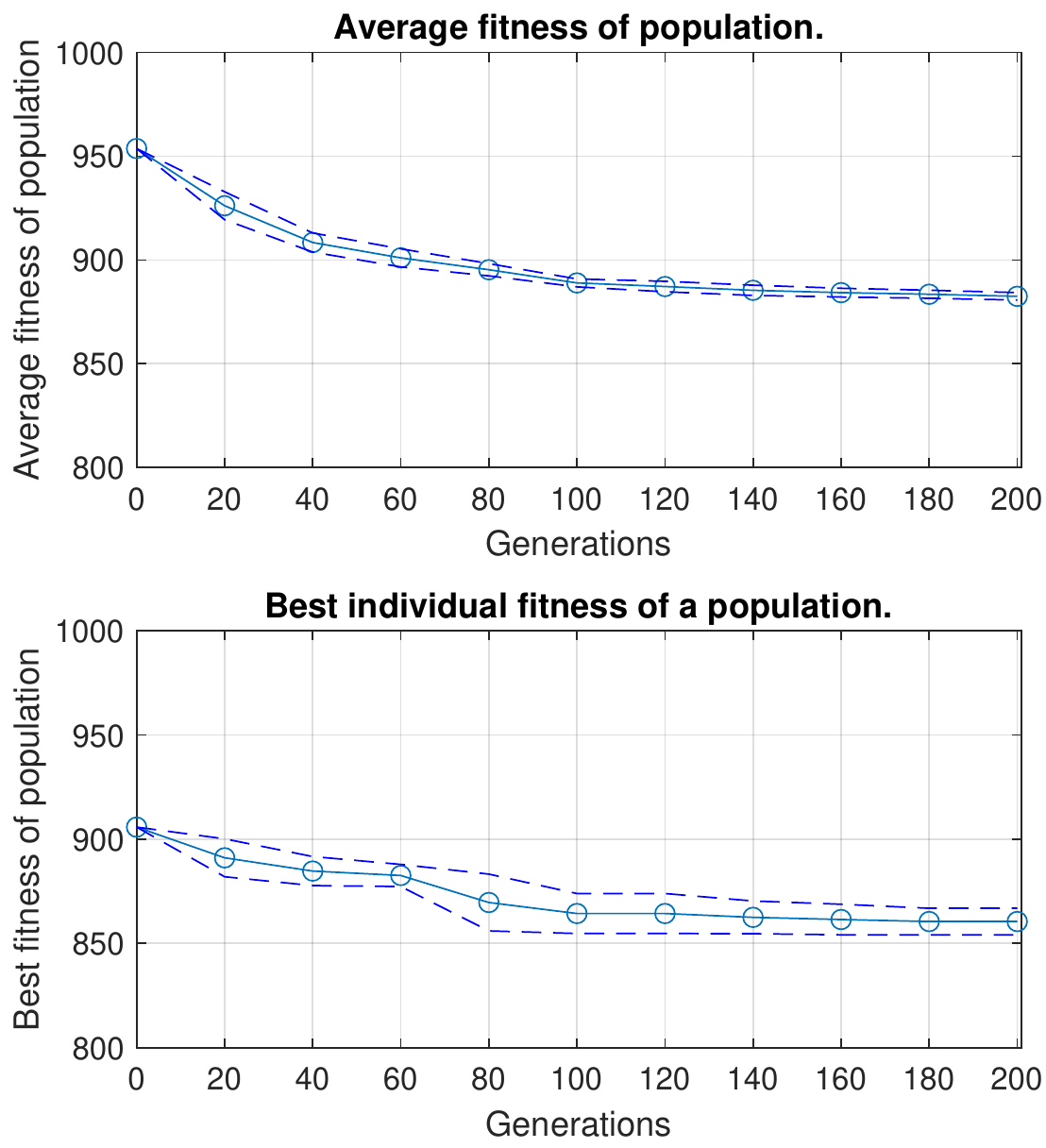}
    \caption{Average and confidence levels (95\%) results from 5 runs of GA. (a)~Evolution of average fitness of the population and (b) evolution of the best individual in the population.}
    \label{physiRes1tot}
\end{figure}

The results of the evolution with the GA are depicted in Figs. \ref{physiRes1} (one instance) and \ref{physiRes1tot} (average of 5 runs). Fig. \ref{physiRes1}(a) shows the average fitness of the population (remaining cancer cells agents) is converging to c. 880 agents after 200 generations of the evolution of the population, from c. 950 agents of the initial random population. That is an improvement of c. 7.3\%. Quite similar results are provided by all the tests executed, as illustrated in Fig. \ref{physiRes1tot}(a) of the cumulative results of average and 95\% confidence levels. For the best individual in the population the final fitness seems to converge close to 850 agents, while the initial randomly generated one has a fitness of c. 900, an improvement of 5.5\%, for one example run (Fig. \ref{physiRes1}(b)). Whereas, the cumulative results depicted in Fig. \ref{physiRes1tot}(b), reveals that during the rest of the runs the improvement is slightly smaller.

In order to enable the variation of genome length - and hence more than one types of NP per solution - the mutation operator of the above methodology was incorporated. The mutation can therefore alter a randomly chosen gene allele or add one type of NPs (with randomly chosen parameters) to the simulated treatment. Both cases have the same possibility of occurring, namely 50\%. Note, that as explained before, no matter the amount of types of NPs, 50 worker agents are injected in the simulated treatment. The initial population used for all tests is set the same as for the previous tests (only gene allele mutation) and consisted of solutions with only one type on NPs. Note that the existence of an additional type of NP potentially alters the fitness landscape significantly due to the complex interactions between the different types of NPs with the cancer cell agents.

\begin{figure}[!t]
    \centering
    \includegraphics[width=0.45\textwidth]{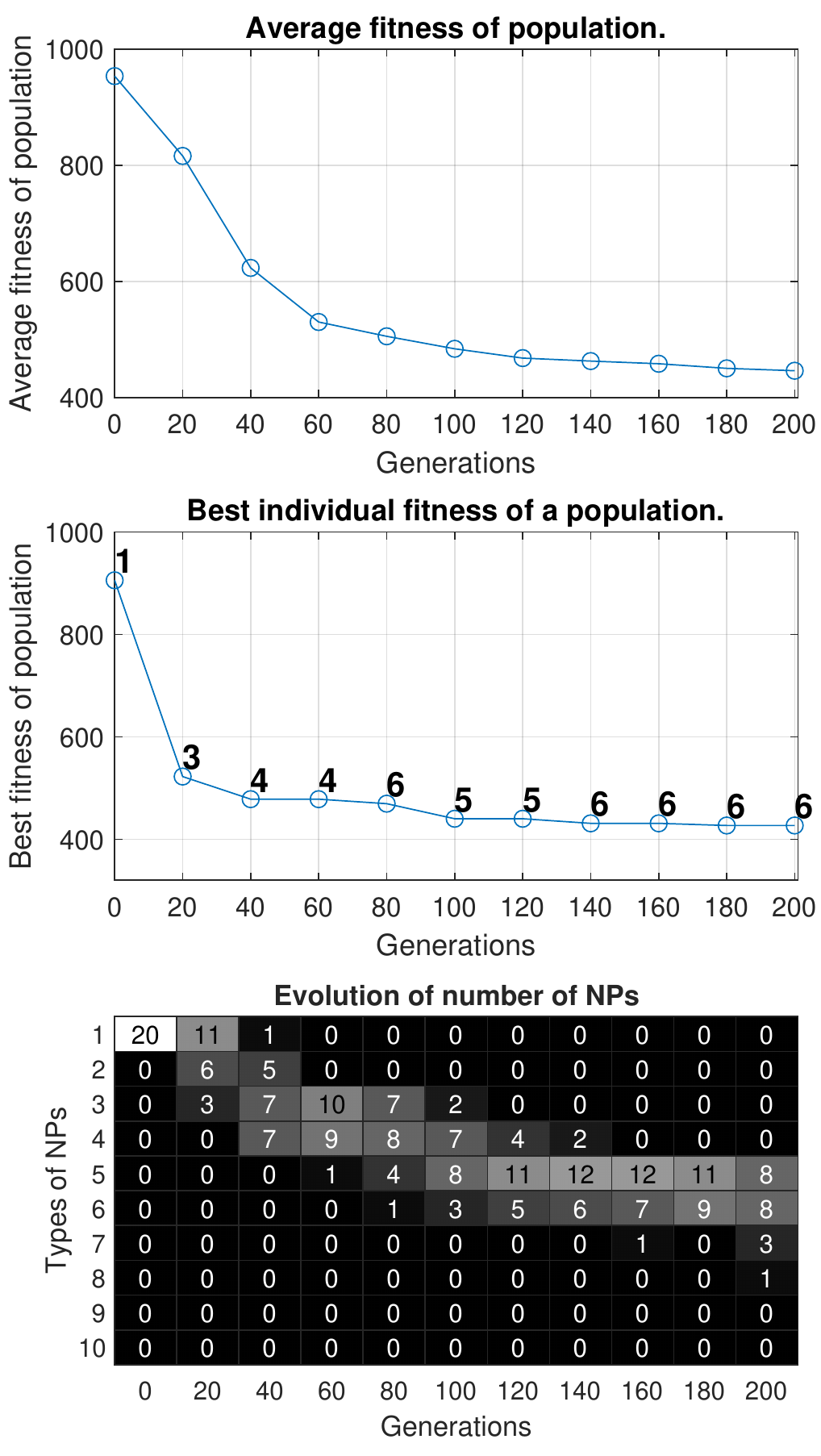}
    \caption{Results from one run of GA with variable-length genome. (a)~Evolution of average fitness of the population, (b) evolution of the best individual in the population (where numbers indicate the number of types NPs) and (c)~composition of population in terms of types of NPs.}
    \label{physiRes2}
\end{figure}



\begin{figure}[!t]
    \centering
    \includegraphics[width=0.45\textwidth]{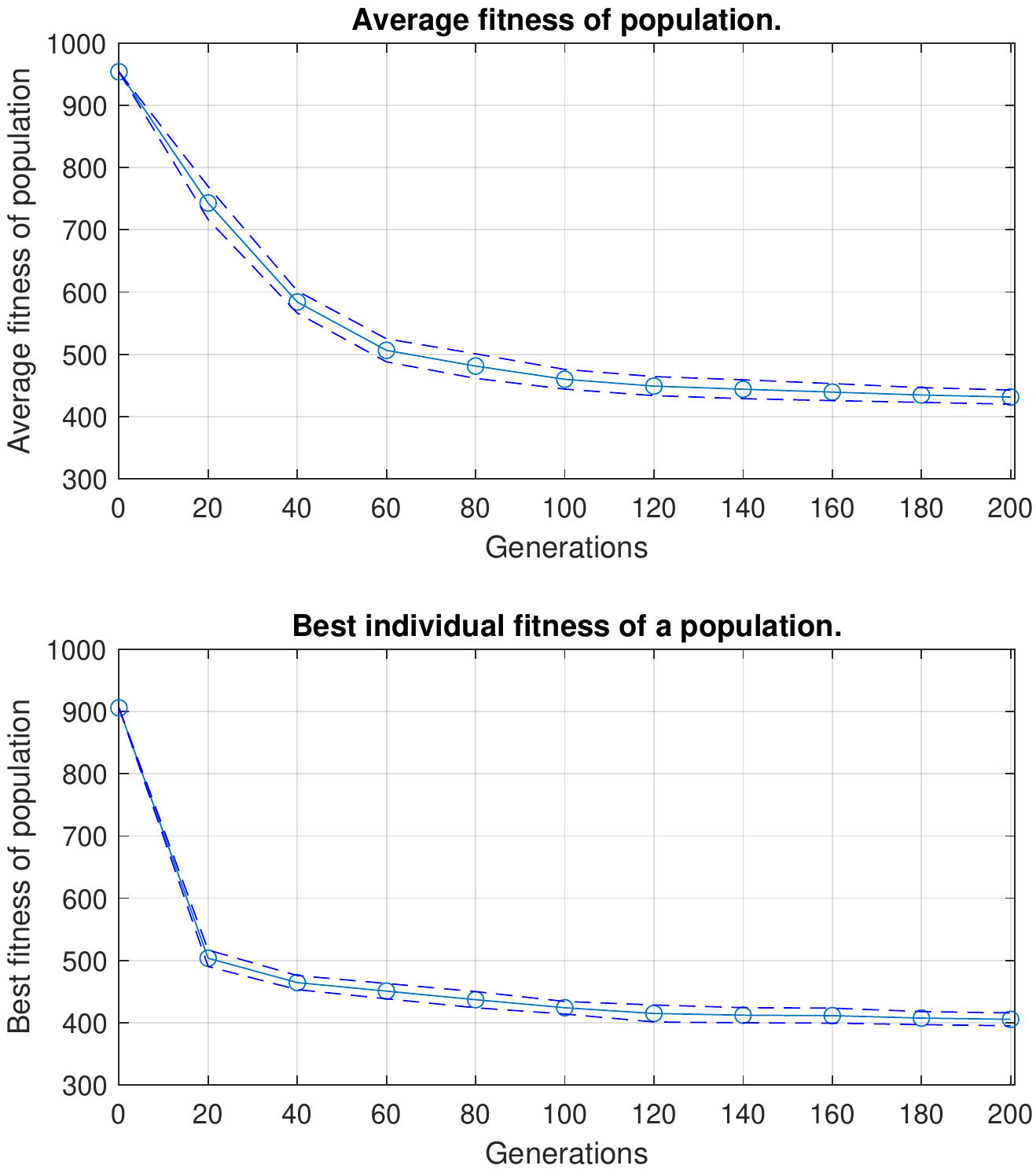}
    \caption{Average and confidence levels (95\%) results from 10 runs of GA with variable-length genome. (a) Evolution of average fitness of the population, (b)~evolution of the best individual in the population.}
    \label{physiRes3tot}
\end{figure}

With the ability to add multiple types of NPs the optimization process reaches better results within the 20 generations as depicted in Figs. \ref{physiRes2} (one instance) and \ref{physiRes3tot} (average of 10 runs). In Fig. \ref{physiRes2}(a), the average fitness of the population is depicted for one instance, which converges to c. 450 agents at the end of the evolution, while the same metric of the initial random population is c. 950 agents. Thus, the alternation in the amount of types of NPs injected results in an improvement of c. 52.6\%. Investigating the cumulative results of average and 95\% confidence levels of all the runs that are portrayed in Fig. \ref{physiRes3tot}(a), there is an evident further improvement of the average fitness of the population (average of 431 agents). Moreover, there is an improvement of 52.2\% when comparing the final fitness of the best individual discovered during one run (approximately 430 agents) with the randomly generated initial one (c. 900 identical in all runs), as outlined in Fig. \ref{physiRes2}(b). In Fig. \ref{physiRes3tot}(b) where all the results of 10 runs are considered, it can be seen that during the rest of the runs the improvement is similar and slightly better (average of 405 agents).

\begin{figure}[!t]
    \centering
    \includegraphics[width=0.45\textwidth]{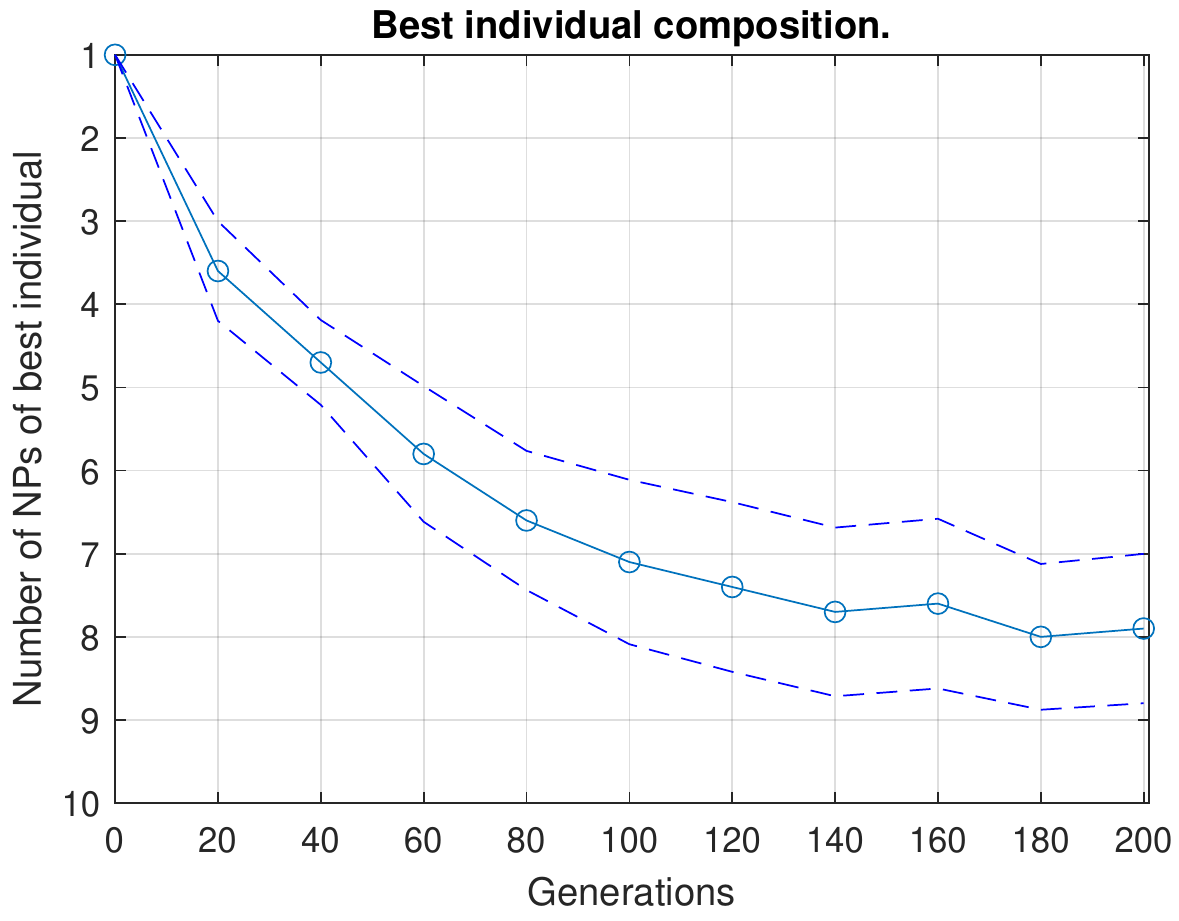}
    \caption{Average and confidence levels (95\%) results from 10 runs of GA with variable-length genome for the composition of the best solution.}
    \label{physiRes3best}
\end{figure}

Despite the fact that the types of NPs can reach the maximum of 10, the composition of the best solutions usually converge to a lower complexity, for instance around 6 types of NPs, as illustrated in Fig. \ref{physiRes2}(b) and \ref{physiRes2}(c). Nonetheless, taking into consideration the average of all 10 tests, solutions converge to 8 types (as illustrated in Fig. \ref{physiRes3best}). Following behaviour observed above in the NK model, as shown in Fig.  \ref{physiRes2}(b), evolution typically adds genes (types of NPs) quite fast, finding the best composition of types of NPs later with the overall optimization continuing until it finds an optimum. Similar behaviour is observed in all instances (Fig. \ref{physiRes3best}).

\begin{figure}[!t]
    \centering
    \includegraphics[width=0.45\textwidth]{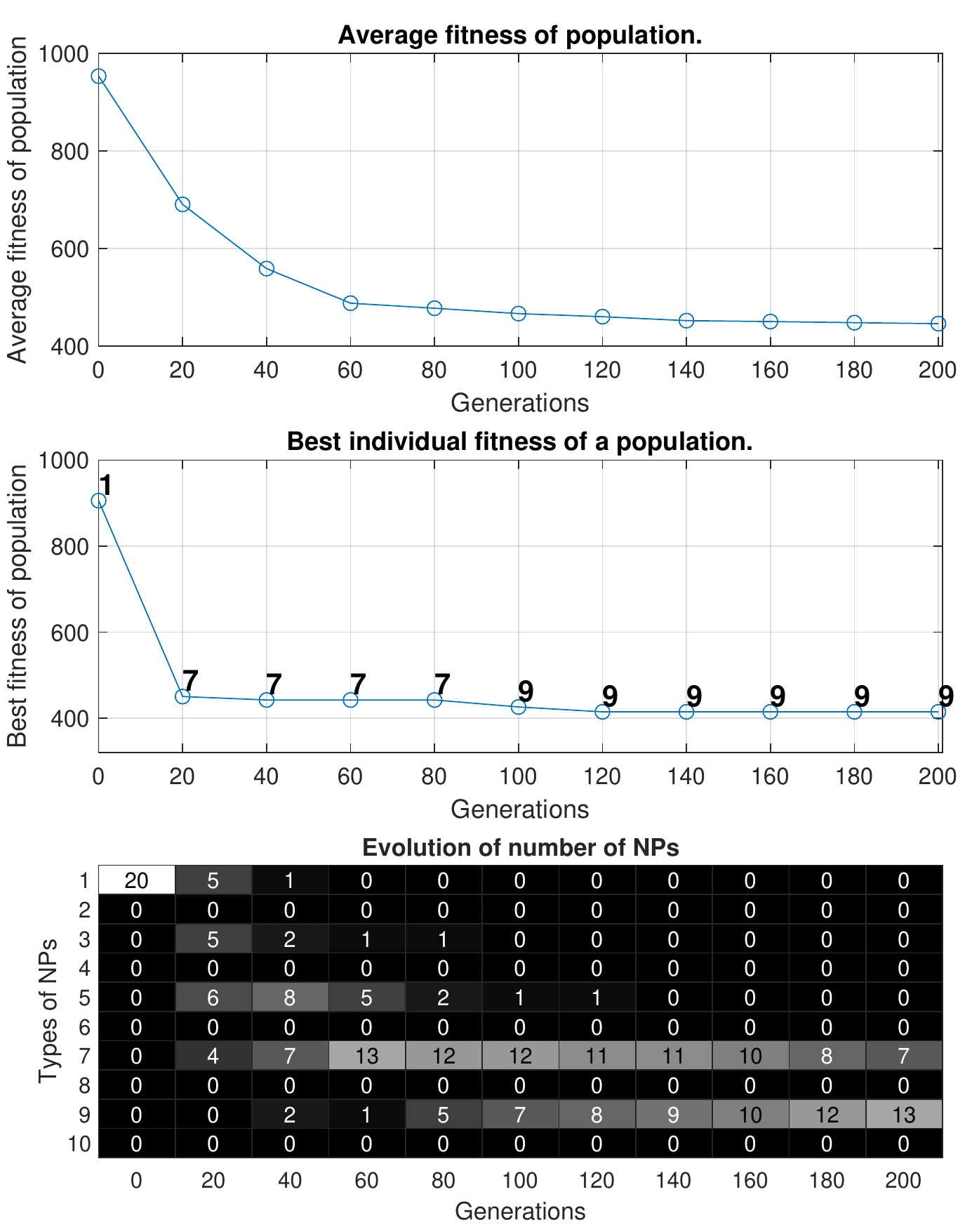}
    \caption{Results from one run of GA with variable-length genome adding two types of NPs. (a)~Evolution of average fitness of the population, (b) evolution of the best individual in the population (where numbers indicate the number of types NPs) and (c)~composition of population in terms of types of NPs.}
    \label{physiResAdd2}
\end{figure}

\begin{figure}[!t]
    \centering
    \includegraphics[width=0.45\textwidth]{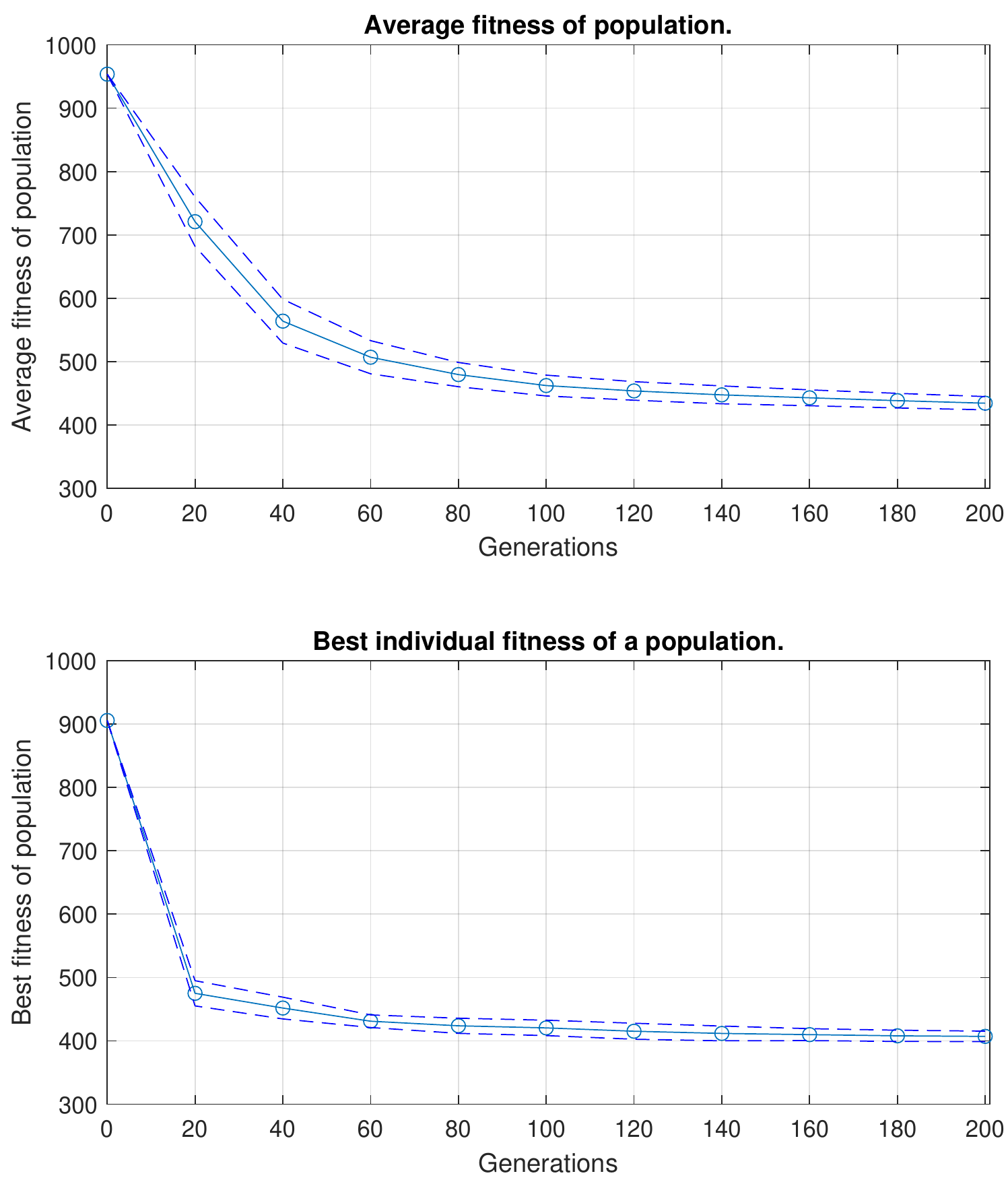}
    \caption{Average and confidence levels (95\%) results from 10 runs of GA with variable-length genome adding two types of NPs. (a) Evolution of average fitness of the population, (b)~evolution of the best individual in the population.}
    \label{physiResTotL2}
\end{figure}

The same procedure as in previous experiment was utilized with mutation adding two new types of NPs. This process reaches similar results to the addition of one type of NP per mutation event, as observed in Figs. \ref{physiResAdd2} (one instance) and \ref{physiResTotL2} (average of 10 runs). The average fitness of the population for one instance, portrayed in Fig. \ref{physiResAdd2}(a), converges to c. 445 agents. That implies an improvement of c. 53.1\%, compared with the same metric of the initial random population is c. 950 agents. When considering the cumulative results of 10 runs, outlined in Fig. \ref{physiResTotL2}(a), it is evident that a further improvement of the average fitness of the population occurs (average of 434 agents). Also, the final fitness of the best individual produced during one run (approximately 415 agents) compared with the randomly generated initial one (c. 900 agents) is improved by 53.9\%, as given in Fig. \ref{physiResAdd2}(b). In Fig. \ref{physiResTotL2}(b) where all the results of 10 runs were considered, it can be established that during the rest of the runs the improvement is similar and slightly better (average of 407 agents).

\begin{figure}[!t]
    \centering
    \includegraphics[width=0.45\textwidth]{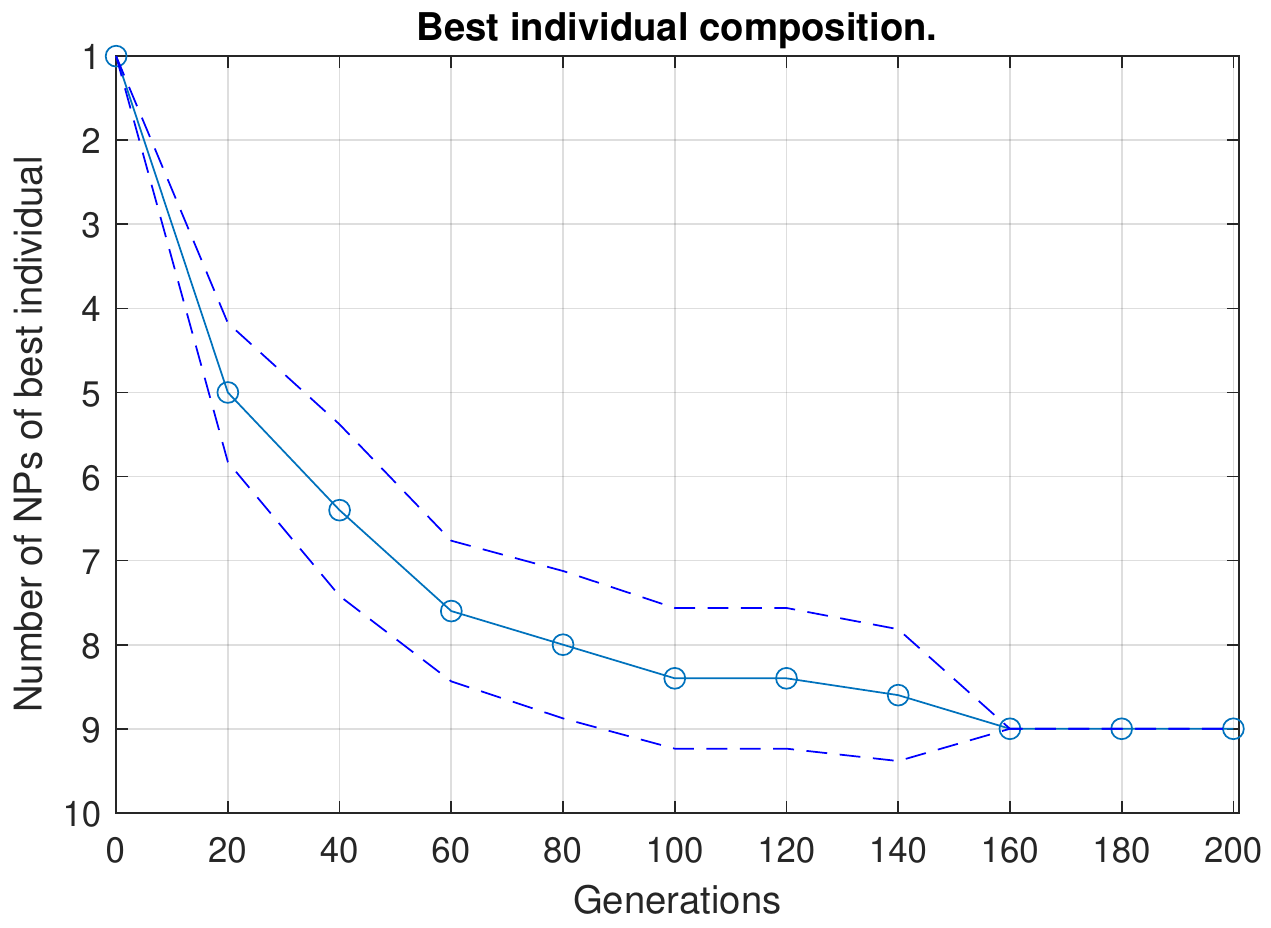}
    \caption{Average and confidence levels (95\%) results from 10 runs of GA with variable-length genome adding two types of NPs for the composition of the best solution.}
    \label{physiResBestL2}
\end{figure}

Here, in contrast with adding one type of NP per mutation event, the best solutions converge to the maximum available composition, namely 9 types, as illustrated in Fig. \ref{physiResBestL2}. In some of the runs the highest amount of NPs is reached by the 100th generation (earlier than the previous case, i.e. the 120th generation), while in all runs the composition of best solutions converges to 9 by the 160th generation.

Finally, note that as there was no significant difference of the genome length in the results presented while investigating the \textit{NK} model and the possibility of decreasing the genome length (in addition to increasing the genome length), this technique was not tested in the computationally expensive simulator.

\section{Conclusion}

This paper has considered the effects of fitness landscape ruggedness on the evolution of genome length under different conditions. Using a simple, abstract model it has been shown that increases in genome length due to the addition of randomly created novel sequences which have an immediate effect upon fitness is the norm. That is, fitness landscape ruggedness does not hinder genome growth and can actually promote it as the window of opportunity for a randomly created sequence to have a beneficial -- or at least neutral -- effect on fitness increases due to the typical (low) height of peaks in such landscapes. Note no explicit functional benefit from increased genome length was included in the model and hence the growth seen was as a consequence of the underlying dynamics of evolution on fitness landscapes of varying degrees of ruggedness only.

Increasing the genome length in a real-world problem was, then, investigated. Specifically, the optimization of the design of NPs that comprise the drug delivery system on a cancer tumour through a physic-based cell simulator was used. Despite the fact that no indication of the best composition of the treatment (in terms of the amount of types of NPs) was included in the model, the results of the evolution seem to converge to treatments with eight types of NPs, when adding one type of NPs per mutation event, and to slightly more complex treatments with nine types, when adding two types of NPs. As deduced by the examination of the abstract \textit{NK} model (after \cite{harvey1992species}), the gradual growth through small step increases in genome length appears more appropriate here. That is, whilst the fitness of the solutions found is quite similar, the higher complexity of NP-based cancer treatment drug delivery systems is harder to produce, will probably prove to be more toxic, and has the greater potential for unintended consequences when used in vivo.

As no significant difference arose from having both addition and deletion in the genome length while testing the abstract model, this was not tested on the computationally expensive simulator. As an aspect of future work, different growth and deletion schemes will be explored, such as those which self-adapt over time (after \cite{bull2005coevolutionary}), along with recombination operators.

\section{Acknowledgments}

This work was supported by the European Research Council under the European Union's Horizon 2020 research and innovation programme under grant agreement No. 800983.

\end{document}